\documentclass[journal]{IEEEtran}
\usepackage{cite}
\ifCLASSINFOpdf
   \usepackage[pdftex]{graphicx}
   \graphicspath{{../pdf/}{../jpeg/}}
   \DeclareGraphicsExtensions{.pdf,.jpeg,.png}
\else
   \usepackage[dvips]{graphicx}
   \graphicspath{{../eps/}}
   \DeclareGraphicsExtensions{.eps}
\fi
\usepackage{amsmath}
\usepackage{amssymb}
\usepackage{algorithm}
\usepackage{booktabs}
\usepackage{multirow}
\usepackage{cellspace}
\usepackage{stackengine}
\usepackage[hidelinks]{hyperref}

\newlength{\continueindent}
\usepackage{graphicx}
\usepackage{subcaption}
\makeatother
\usepackage{array}

\usepackage{etoolbox}
\usepackage{bm}
\usepackage{amsthm}%

\newcommand{\MyMapTemplatePrefixc}[4]{\expandafter#1\csname#3#4\endcsname{#2{#4}}} % it remembles a template: \#3#4 --> #2{#4}
\forcsvlist{\MyMapTemplatePrefixc {\def} {\mathcal}{c}} {A,B,C,D,E,F,G,H,I,J,K,L,M,N,O,P,Q,R,S,T,U,V,W,X,Y,Z}  % e.g., \cA --> \mathcal{A}

\newcommand{\MyMapTemplatePrefixtb}[5]{\expandafter#1\csname#4#5\endcsname{#2{#3{#5}}}} % it remembles a template: \#3#4 --> #2{#4}
\forcsvlist{\MyMapTemplatePrefixtb {\def} {\tilde}{\mathbf}{t}} {A,B,C,D,E,F,G,H,I,J,K,L,M,N,O,P,Q,R,S,T,U,V,W,X,Y,Z}  % e.g., \tA --> \tilde{A}
\forcsvlist{\MyMapTemplatePrefixtb {\def} {\tilde}{\mathbf}{t}} {0,1,a,b,c,d,e,f,g,h,i,j,k,l,m,n,o,p,q,r,s,u,v,w,x,y,z}  % e.g., \ta --> \tilde{a}

\newcommand{\MyMapTemplateNoPrefix}[3]{\expandafter#1\csname#3\endcsname{#2{#3}}}
\forcsvlist{\MyMapTemplateNoPrefix {\def} {\mathbf} } {0,1,a,b,c,d,e, f, g, h, i, j, k, l, m, n, o, p, q, r, u, v, w, x, y, z} % e.g., \a --> \mathbf{a}
\forcsvlist{\MyMapTemplateNoPrefix {\def} {\mathbf} } {A,B,C,D,E,F,G,H,I,J,K,L,M,N,O,P,Q,R,S,T,U,V,W,X,Y,Z}  % e.g., \A --> \mathcal{A}

\def\etal{\emph{et al.}}
\def\ie{\emph{i.e.}}
\def\eg{\emph{e.g.}}

\usepackage{booktabs} % Nice tables
\usepackage[table,dvipsnames]{xcolor}
\definecolor{rowblue}{RGB}{220,230,240}

\begin{document}
%
% paper title
% Titles are generally capitalized except for words such as a, an, and, as,
% at, but, by, for, in, nor, of, on, or, the, to and up, which are usually
% not capitalized unless they are the first or last word of the title.
% Linebreaks \\ can be used within to get better formatting as desired.
% Do not put math or special symbols in the title.
%\title{Modality-Specific and Modality-Grouped Sample Cooperation for Unsupervised Domain Adaptation and Unsupervised Domain Adaptation with (and without?) Privileged Information}
\title{An Agentic Framework for Quality-Aware and Scenario-Adaptive Image Compression}
%
%

% author names and IEEE memberships
% note positions of commas and nonbreaking spaces ( ~ ) LaTeX will not break
% a structure at a ~ so this keeps an author's name from being broken across
% two lines.
% use \thanks{} to gain access to the first footnote area
% a separate \thanks must be used for each paragraph as LaTeX2e's \thanks
% was not built to handle multiple paragraphs
%

\author{
Zhenghao Chen, Lebin Zhou, Zicong Chen, Naifu Xue, Jiahao Li, Bin Li, Dong Xu, \textit{Fellow, IEEE,} and Yan Lu
\thanks{Zhenghao Chen and Zicong Chen are with The University of Newcastle. (\textbf{E-mail:} zhenghao.chen@newcastle.edu.au, csc20001218@gmail.com)}
\thanks{Lebin Zhou is with Santa Clara University. (\textbf{E-mail:}  lzhou@scu.edu)}
\thanks{Naifu Xue is with The Communication University of China (\textbf{E-mail:} aaronxuenf@cuc.edu.cn)}
\thanks{Jiahao Li, Bin Li and Yan Lu are with Microsoft Research Asia. (\textbf{E-mail:} \{li.jiahao, libin, yanlu\}@microsoft.com).}
\thanks{Dong Xu is with The University of Hong Kong (\textbf{E-mail:} dongxu@hku.hk).}
\thanks{
This work was initiated and largely completed during Zhenghao Chen's participation in the Microsoft Research Asia  StarTrack Scholars Program.}
\thanks{Bin Li and Yan Lu are the Corresponding Authors.
}
\thanks{
Zhenghao Chen, Lebin Zhou and Zicong Chen contributed equally.}
}
% <-this % stops a space

% \thanks{
% * Dong Xu is corresponding author.}% <-this % stops a space
% \thanks{Manuscript received XX XX, 2020}

% note the % following the last \IEEEmembership and also \thanks - 
% these prevent an unwanted space from occurring between the last author name
% and the end of the author line. i.e., if you had this:
% 
% \author{....lastname \thanks{...} \thanks{...} }
%                     ^------------^------------^----Do not want these spaces!
%
% a space would be appended to the last name and could cause every name on that
% line to be shifted left slightly. This is one of those "LaTeX things". For
% instance, "\textbf{A} \textbf{B}" will typeset as "A B" not "AB". To get
% "AB" then you have to do: "\textbf{A}\textbf{B}"
% \thanks is no different in this regard, so shield the last } of each \thanks
% that ends a line with a % and do not let a space in before the next \thanks.
% Spaces after \IEEEmembership other than the last one are OK (and needed) as
% you are supposed to have spaces between the names. For what it is worth,
% this is a minor point as most people would not even notice if the said evil
% space somehow managed to creep in.

% The paper headers
% \markboth{IEEE TRANSACTIONS ON IMAGE PROCESSING, 2026}%
% {Shell \MakeLowercase{\textit{et al.}}: Bare Demo of IEEEtran.cls for IEEE Journals}
\markboth{Technical Report}%
{Shell \MakeLowercase{\textit{et al.}}: Bare Demo of IEEEtran.cls for IEEE Journals}
% The only time the second header will appear is for the odd numbered pages
% after the title page when using the twoside option.
% 
% *** Note that you probably will NOT want to include the author's ***
% *** name in the headers of peer review papers.                   ***
% You can use \ifCLASSOPTIONpeerreview for conditional compilation here if
% you desire.

% If you want to put a publisher's ID mark on the page you can do it like
% this:
%\IEEEpubid{0000--0000/00\$00.00~\copyright~2015 IEEE}
% Remember, if you use this you must call \IEEEpubidadjcol in the second
% column for its text to clear the IEEEpubid mark.

% use for special paper notices
%\IEEEspecialpapernotice{(Invited Paper)}

% make the title area
\maketitle

% As a general rule, do not put math, special symbols or citations
% in the abstract or keywords.
\begin{abstract}
Despite decades of progress, image codecs remain difficult to adapt to diverse practical compression scenarios for two main reasons: First, most image codecs are optimized for a specific objective and require substantial redevelopment when the compression goal changes. Second, they typically assume clean inputs, whereas sensor-captured images often contain degradations that alter rate--quality behavior. To address these challenges, we propose Practical Image Coding Agents (PICA), a multi-agent framework that adapts existing codecs to diverse user requirements and potentially degraded inputs.
PICA consists of four collaborative agents: a \underline{\textit{P}}erception agent that interprets user intent and analyzes image content, an \underline{\textit{I}}mprovement agent that jointly considers bitrate and reconstruction quality to guide pre-coding restoration, a \underline{\textit{C}}oding agent that refines quality-control settings from global to regional levels, and an \underline{\textit{A}}fter-coding improvement agent that addresses residual degradations and compression-induced artifacts through restoration guided solely by image quality.
To enable codec-agnostic regional adaptation, we introduce Agentic ROI Coding (AROIC), which treats the codec as a black box and allows the agent to automatically schedule, execute, evaluate, and reflect ROI coding plans without architectural modifications or codec-specific retraining. We integrate these four agents with a Scenario-Guided Quality--Rate (SGQR) evaluation strategy to enable more effective scenario-adaptive compression across heterogeneous codecs.
Comprehensive experiments on clean, noisy, and blurred images across standard, perceptual, and machine-vision compression scenarios demonstrate that PICA consistently enhances the compression performance of both the conventional H.266/VVC codec and the neural Cheng2020 codec, outperforming competing compression methods on the Kodak, CLIC2020, and HEVC datasets.
\end{abstract}

% Note that keywords are not normally used for peerreview papers.
\begin{IEEEkeywords}
Image compression, image restoration, agentic systems, perceptual compression, machine-vision compression.
\end{IEEEkeywords}

% For peer review papers, you can put extra information on the cover
% page as needed:
% \ifCLASSOPTIONpeerreview
% \begin{center} \bfseries EDICS Category: 3-BBND \end{center}
% \fi
%
% For peerreview papers, this IEEEtran command inserts a page break and
% creates the second title. It will be ignored for other modes.
\IEEEpeerreviewmaketitle

\section{Introduction}
\IEEEPARstart{I}{mage} compression is a fundamental problem in signal processing, as the rapidly growing volume and resolution of visual data impose substantial costs on storage, transmission bandwidth, and communication latency. This has been extensively studied for more than three decades, leading to mature hand-engineered image coding standards such as JPEG~\cite{wallace1992jpeg}. Such traditional codecs remain widely used in practical image storage and communication systems. More recent standards, such as H.266/VVC~\cite{bross2021vvc}, also support high-efficiency image compression through intra-frame coding. With rapid advances in deep learning, neural image compression (NIC) methods have achieved substantial progress by jointly optimizing nonlinear transforms and entropy models under end-to-end rate--distortion (RD) objectives~\cite{balle2017end,balle2018variational,minnen2018joint,lee2019context,mentzer2020high,chen2022exploiting,Chen_2022_CVPR,chen2023neural}. Representative methods such as Cheng2020~\cite{cheng2020learned} have demonstrated RD performance competitive with or superior to those traditional codecs on the standard compression benchmarks.

Despite this tremendous success, most widely deployed codecs are designed primarily for generic pixel-fidelity reconstruction, with their RD performance conventionally optimized and evaluated using distortion metrics such as PSNR, a compression scenario referred to as \emph{standard compression}. However, modern visual-data ecosystems increasingly involve heterogeneous consumers and objectives with distinct compression intents. For example, some compressed images are primarily expected to preserve visual realism and semantic fidelity for human observers~\cite{blau2019rethinking,agustsson2019generative,mentzer2020high,careil2024perco}, a compression scenario referred to as \emph{perceptual compression}. Meanwhile, an increasing proportion of visual data is consumed by downstream machine-learning systems, where compression should preserve task-relevant information and downstream accuracy~\cite{chamain2021end,liu2023icmh,li2024adapticmh}, a compression scenario referred to as \emph{machine-vision compression}. However, prior studies~\cite{mentzer2020high,careil2024perco,liu2023icmh,li2024adapticmh} have shown that commonly used standard codecs can become suboptimal in these specialized compression scenarios.

Developing dedicated codecs for these specialized scenarios provides an intuitive solution. Building on the advantages of NIC, several scenario-specific compression methods have been proposed. For example, machine-vision compression approaches~\cite{liu2023icmh,li2024adapticmh} introduce task-aware compression modules to produce latent representations that preserve information relevant to downstream vision tasks. Meanwhile, perceptual NIC methods~\cite{agustsson2019generative,mentzer2020high} combine powerful generative models with human-perception-oriented objectives, such as LPIPS~\cite{zhang2018lpips}, to reconstruct visually plausible content. Though these specialized designs achieve substantial improvements in their respective scenarios, they are typically tightly coupled with extensive scenario-specific modules and objectives, thereby requiring considerable architectural redesign and substantial re-optimization when the compression intent changes. Consequently, they exhibit limited generalizability across diverse compression scenarios. This limitation motivates our first research question (RQ): \emph{RQ1: Can existing widely-used codecs be rapidly adapted to diverse compression scenarios without architectural redesign or substantial re-optimization?}

Beyond their limited generalizability across compression scenarios, most existing codecs are developed under the assumption of clean input images. In practice, however, acquired images may exhibit degradations (\eg, noise) due to limitations of imaging sensors, such as those in smartphone cameras. These degradations not only compromise visual quality but also introduce undesirable signal variations that may increase the number of bits required for compression~\cite{cheng2022optimizing,ye2023accelir}.
To address this issue, recent studies have jointly optimized restoration and compression or introduced restoration-aware compression modules~\cite{cheng2022optimizing,brummer2023importance,ye2023accelir,cai2024lowlight,chen2024joint,10.1145/3743093.3770957, zafari2026zero}. However, these approaches still rely on dedicated components and degradation-specific optimization. Moreover, they generally assume that the degradation type and restoration objective are known in advance, rather than autonomously determining whether restoration is necessary and how it should be coordinated before and after coding. It motivates our second RQ: \emph{RQ2: Can an existing widely used codec autonomously detect image degradations and adaptively coordinate pre-coding and post-coding restoration without substantial re-development?}

To simultaneously address above two RQs, we propose \textit{Practical Image Coding Agents (PICA)}, a multi-agent framework that adapts existing image codecs to diverse degraded inputs and compression scenarios through online and goal-oriented optimization. PICA consists of four collaborative agents: a \underline{P}erception Agent, an \underline{I}mprovement Agent, a \underline{C}oding Agent, and an \underline{A}fter-Coding Improvement Agent. Specifically, the Perception Agent interprets the user-specified compression intent while analyzing the input image, including its visual content, quality condition, and regions of interest, thereby providing structured guidance for subsequent agents. 

To address \textit{RQ1}, we introduce a Coding Agent with a global-to-regional coding quality refinement scheme.
Guided by the compression intent summarized by the Perception Agent, it first performs Global \emph{Quality Control (QC)} Search, applying estimated QC to the target codec and iteratively refining them based on the measured bitrate and reconstruction quality until the user-specified requirements are satisfied.
Starting from this feasible global operating point, we further refine the coding quality of individual regions, motivated by prior studies showing that different compression scenarios assign different importance to image regions. For example, machine-vision compression often requires stronger preservation of task-relevant objects than background regions~\cite{li2024adapticmh,liu2023icmh}. Yet, most existing compression methods~\cite{gao2025compx} rely on codec-specific ROI coding mechanisms, requiring the underlying codec to be explicitly designed or trained with dedicated ROI functionality and thus limiting their generalizability to heterogeneous codecs without native ROI support.

To overcome this limitation, we introduce Agentic ROI Coding (AROIC), which treats the underlying codec as a black box and follows a scheduling–execution–reflection loop. Specifically, the agent schedules and executes candidate coding plans that adjust the QC settings of the ROIs and background and, when beneficial, apply background downsampling. Each region is independently encoded using the target codec with its assigned QC setting, and the decoded regions are assembled into a complete reconstruction. The agent then evaluates the compression results and uses the evaluation feedback to refine subsequent coding plans while retaining the best-performing candidate. By relying on the codec's native QC interface, AROIC enables ROI-aware adaptation across heterogeneous codecs without codec-specific redesign or retraining.

Moreover, to address \textit{RQ2}, we introduce the {Improvement Agent} and the {After-Coding Improvement Agent}. 
First, the Improvement Agent performs quality--rate (QR)-aware pre-coding restoration to mitigate input degradations and improve the overall QR trade-off. Unlike existing restoration agents~\cite{zhu2025intelligent,zuo20254kagent}, which focus on correcting all detected degradations, our agent jointly evaluates the coding rate and reconstruction quality resulting from each candidate restoration operation. It retains only operations that improve the QR trade-off under the specified compression objective.
Similarly, the After-Coding Improvement Agent performs quality-aware post-coding restoration to mitigate residual degradations and potential compression-induced artifacts in the decoded image. This stage focuses on degradations that primarily affect reconstruction quality and adaptively applies restoration only when it improves the quality of the final decoded image.

% SGQR
To align the collaborative agents with the intended compression scenario, we introduce a Scenario-Guided Quality–Rate (SGQR) evaluation strategy for the Improvement and Coding Agents. Specifically, SGQR evaluates each candidate improvement or coding plan using the bitrate and reconstructed image obtained after compression. It assesses reconstruction quality using scenario-specific metrics and selects the plan that offers the best QR trade-off under the user's compression requirements.
Overall, our PICA integrates four collaborative agents with SGQR evaluation to adapt heterogeneous existing codecs to diverse compression scenarios without codec-specific redevelopment. We conduct extensive experiments on clean, noisy, and blurred images across standard, perceptual, and machine-vision compression scenarios. Results on the Kodak~\cite{franzen1999kodak}, CLIC2020~\cite{toderici2020clic}, and HEVC~\cite{sullivan2012overview} datasets demonstrate that PICA consistently improves the compression performance of both H.266/VVC~\cite{bross2021vvc} and Cheng2020~\cite{cheng2020compression}.

Our main contributions can be summarized as follows:
\begin{itemize}
    \item \textbf{Practical Image Coding Agents.} We propose a multi-agent framework that adapts existing image codecs to diverse compression scenarios and potentially degraded inputs through SGQR-guided optimization.

    \item \textbf{Quality--Rate--Aware Improvement Agents.} We introduce a pre-coding improvement agent that jointly considers bitrate and reconstruction quality to address input degradations, and a post-coding improvement agent to mitigate residual and compression-induced degradations.

    \item \textbf{Codec--Agnostic Coding Agent.} We introduce a Coding Agent with global-to-regional coding quality refinement together with AROIC, which treats the underlying codec as a black box and enables ROI-aware adaptation across heterogeneous codecs without codec-specific redesign.
\end{itemize}

\section{Related Works}
\subsection{Image Compression}
Image compression methods broadly fall into conventional codecs and NIC. Conventional standards include JPEG~\cite{wallace1992jpeg}, JPEG2000~\cite{skodras2001jpeg2000}, and intra-frame coding mode in H.265/HEVC~\cite{sullivan2012overview} and H.266/VVC~\cite{bross2021vvc}. These standard codecs form the modern image compression systems. Driven by expressive neural architectures and end-to-end optimization, NIC has advanced substantially through hyperpriors~\cite{balle2018variational}, autoregressive context models~\cite{minnen2018joint}, and Gaussian mixture entropy models with attention~\cite{cheng2020learned}, the latter achieving RD performance competitive with those standard codecs.

Beyond conventional compression that prioritizes pixel-wise reconstruction fidelity, growing attention has been devoted to alternative compression objectives. In this work, we focus on two challenging scenarios: perceptual compression~\cite{blau2019rethinking,agustsson2019generative,mentzer2020high,careil2024perco} and machine-vision compression~\cite{chamain2021end,liu2023icmh,li2024adapticmh}.
Existing approaches address these objectives through specialized training strategies and architectural components. For example, HiFiC~\cite{mentzer2020high} combines adversarial training and perceptual optimizations to produce visually realistic reconstructions. Adapt-ICMH~\cite{li2024adapticmh} optimizes a spatial-frequency modulation adapters to suppress task-irrelevant redundancy. Although effective for the specific scenarios, these approaches require objective-specific training or additional trainable components. In contrast, PICA is a training-free framework that adapts existing black-box codecs at inference time without codec re-optimization or scenario-specific module redesign. Through global-to-regional quality control, it allocates bits according to perceptual or task-specific priorities.

\begin{figure*}[t!]
    \centering
    \includegraphics[width=\textwidth]{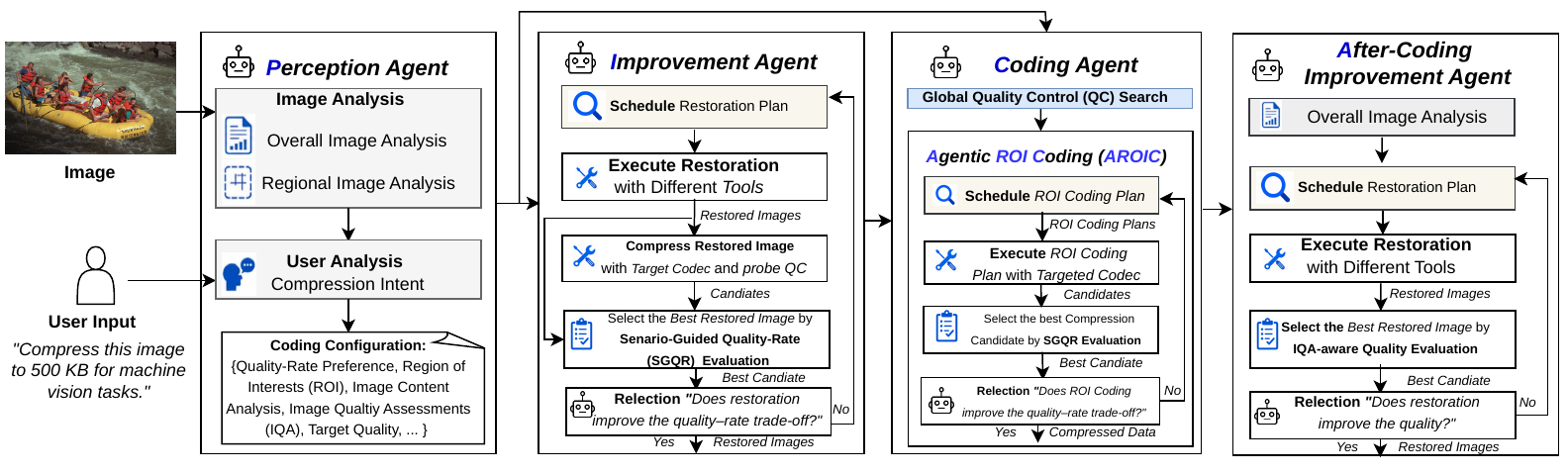}
    \caption{Overview of the \textbf{P}ractical \textbf{I}mage \textbf{C}ompression \textbf{A}gent (\textbf{PICA}) framework, consisting of four collaborative agents: the \underline{P}erception agent, the \underline{I}mprovement agent, the \underline{C}oding agent, and the \underline{A}fter-coding improvement agent.}
    \label{fig:framework}
\end{figure*}

\subsection{Quality-Aware Image Compression}
Most existing image compression methods treat the observed input as the reconstruction target, without explicitly accounting for acquisition degradations. In practice, however, images may contain noise, blur, or other degradations~\cite{cheng2022optimizing,brummer2023importance,cai2024lowlight,10.1145/3743093.3770957,zafari2026zero}. These degradations not only compromise visual quality but may also lead codecs to allocate additional bits to preserving undesirable signal variations rather than prioritizing the underlying clean signal~\cite{cheng2022optimizing,brummer2023importance}.

A common approach to addressing this issue is to jointly optimize compression and restoration~\cite{cheng2022optimizing,brummer2023importance,ye2023accelir,cai2024lowlight,10.1145/3743093.3770957}. For example, Cheng~\etal~\cite{cheng2022optimizing} introduce a two-branch, weight-sharing architecture with feature denoisers for joint denoising and compression, while Wan~\etal~\cite{10.1145/3743093.3770957} explore joint compression and deblurring through blur-aware guidance. To avoid retraining the entire codec, Chen~\etal~\cite{chen2024joint} freeze a pre-trained codec and train additional latent refinement and prompt-generation modules for joint decoding and denoising. Overall, integrating restoration into these compression frameworks requires training either the full model or additional architectural components.
In contrast, PICA coordinates off-the-shelf restoration tools and black-box codecs without additional training. At inference time, it diagnoses input degradations and evaluates candidate restoration tools to select suitable operations. Before coding, it addresses input degradations by jointly considering bitrate and reconstruction quality. After decoding, it mitigates residual and compression-induced degradations.

\subsection{Intelligent Agents for Image Processing}
Intelligent agents are systems that autonomously perceive and act upon their environments to pursue specified goals~\cite{franklin1997agent}. Recent advances in Large Language Models (LLMs) have enabled frameworks such as ReAct~\cite{yao2023react}, which interleaves reasoning and actions to guide decision-making through environmental feedback. Meanwhile, Vision–Language Models (VLMs)~\cite{openai2023gpt4,bai2025qwen25vl, bai2025qwen3} provide visual understanding capabilities that support perception and assessment in image-processing agents. For example, You~\etal~\cite{you2024depicting} introduce DepictQA, which analyzes image content and distortions to provide descriptive quality assessments beyond scalar scores.

Building on visual assessment and language-based reasoning, recent studies coordinate specialized restoration tools through agentic planning and execution~\cite{chen2024restoreagent,jiang2025multiagent,lin2025jarvisir,zhu2025intelligent,zuo20254kagent}. Zhu~\etal~\cite{zhu2025intelligent} propose AgenticIR to address complex image degradations through perception, scheduling, execution, reflection, and rescheduling. Zuo~\etal~\cite{zuo20254kagent} introduce 4KAgent for image restoration and 4K super-resolution, combining perception-guided planning with quality-driven tool selection and an iterative execution--reflection--rollback process. These works demonstrate the potential of integrating visual assessment, planning, and feedback to adaptively coordinate image restoration and enhancement tools.

For image compression, Gao~\etal~\cite{gao2025compx} propose Comp-X, which integrates an LLM agent with a neural codec specifically designed and trained to work with the agent. Although Comp-X supports multiple compression objectives and ROI coding, these capabilities rely on trained task-specific adapters and internal codec mechanisms, including spatial quality maps and group masks.
Its agentic control therefore remains coupled to this specialized codec, leaving the adaptation of existing codecs without redesign or retraining (\ie, RQ1) unaddressed. In contrast, PICA treats ready-to-use conventional and neural codecs as black boxes, enabling scenario-adaptive compression without codec-specific architectural modifications, task-specific adapters, or retraining. Moreover, Comp-X's workflow does not explicitly consider or address acquisition degradations (\ie, RQ2), whereas PICA considers the effects of degradations on both rate and quality to guide restoration before encoding or after decoding.

\section{Methodology}

\subsection{Preliminaries and Formulation}
\label{sec:formulation}
\label{sec:pre}
\noindent\textbf{Quality-Aware Image Compression.} Following prior work on quality-aware image coding~\cite{cheng2022optimizing}, we let $\mathbf{x}^{\star}\in[0,1]^{H\times W\times 3}$ denote an ideal, degradation-free latent image. In practical compression scenarios, the observed input may already contain one or multiple degradations:
\begin{equation}
    \mathbf{x}=\mathcal{G}_{\delta}(\mathbf{x}^{\star}),
    \label{eq:degraded_input}
\end{equation}
where $\mathcal{G}_{\delta}$ denotes an unknown degradation process parameterized by $\delta$ such as noise or blur. The clean-input case is included by letting $\mathcal{G}_{\delta}$ be the identity mapping. During compression, the codec observes only raw image $\mathbf{x}$, neither $\mathbf{x}^{\star}$ nor $\delta$ is assumed to be available.

\noindent\textbf{QC-Controllable Black-Box Codec.}
PICA supports heterogeneous off-the-shelf codecs without accessing or modifying their internal architectures. We use a unified QC factor $q$ to abstract codec-specific quality-control parameters, such as the quantization parameter (QP) in H.266/VVC. Given an input image $\mathbf{z}$ and a QC factor $q$, we define a codec query as
\begin{equation}
\bigl(\mathbf{b},\widetilde{\mathbf{z}}\bigr)
=
\mathcal{C}(\mathbf{z};q),
\label{eq:blackbox_codec}
\end{equation}
where $\mathbf{b}$ and $\widetilde{\mathbf{z}}$ denote the encoded bitstream and reconstructed image, respectively.
We treat $\mathcal{C}$ as a \emph{black-box codec}, without knowing knowledge of its internal architecture or coding procedure, but only assumes that the QC factor can be adjusted and its corresponding rate--quality behavior can be observed. For example, increasing QP in H.266/VVC produces stronger compression with lower reconstruction quality and bitrate.

\begin{figure*}[t!]
    \centering
    \includegraphics[width=\textwidth]{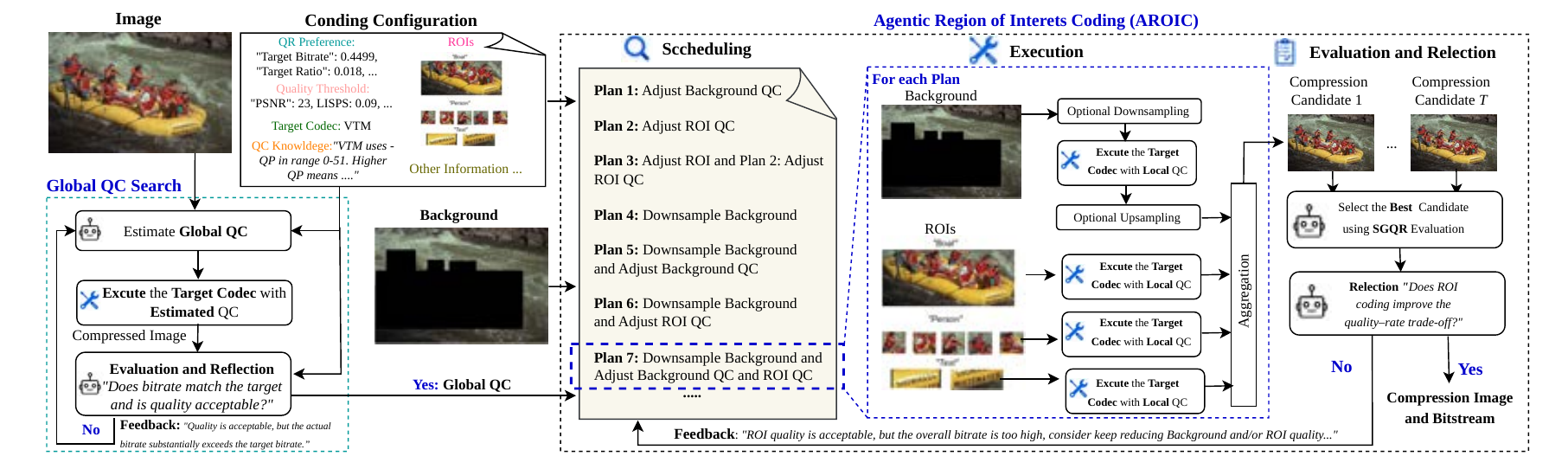}
    \caption{The Coding Agent workflow comprises two components: Global QC Search and Agentic ROI Coding (AROIC).}
    \label{fig:coding}
\end{figure*}

\noindent\textbf{Problem Formulation.}
Given an observed image $\mathbf{x}$ that may contain unknown degradations, a QC-controllable black-box codec $\mathcal{C}$, and a user instruction $\mathcal{I}$ specifying the desired compression objective (\ie, scenario), PICA then employs an agentic coding function $\mathcal{F}$ to produce an aggregate coded representation $\mathcal{B}^{\star}$ and reconstructed image $\widetilde{\mathbf{x}}^{\star}$:
\begin{equation}
\begin{aligned}
\bigl(\mathcal{B}^{\star},\widetilde{\mathbf{x}}^{\star}\bigr)
&=
\arg\min_{\mathcal{B},\widetilde{\mathbf{x}}}
\left[
R(\mathcal{B})
+
\lambda_{\mathcal{I}}
Q_{\mathcal{I}}
\bigl(\widetilde{\mathbf{x}},\mathbf{x}^{\star}\bigr)
\right], \\
& \text{s.t.,}\quad\quad
\bigl(\mathcal{B},\widetilde{\mathbf{x}}\bigr)
=
\mathcal{F}\bigl(\mathbf{x},\mathcal{C},\mathcal{I}\bigr).
\end{aligned}
\label{eq:form}
\end{equation}
Here, $\mathcal{B}$ denotes the complete coded representation required for reconstruction, comprising one or more codec-generated bitstreams and any necessary coding metadata.
$R(\mathcal{B})$ measures its total bitrate, while $\lambda_{\mathcal{I}}$ controls the QR trade-off based on the compression requirements, such as the target bitrate.
The instruction-dependent quality term $Q_{\mathcal{I}}(\widetilde{\mathbf{x}},\mathbf{x}^{\star})$ evaluates reconstruction quality relative to $\mathbf{x}^{\star}$, with lower values indicating better quality.
Rather than measuring distortion against the potentially degraded input $\mathbf{x}$, as is typical in conventional image compression, we follow other quality-aware compression~\cite{cheng2022optimizing} and assess reconstruction quality with respect to the latent degradation-free image $\mathbf{x}^{\star}$.
The quality criterion $Q_{\mathcal{I}}$ is determined by the user-specified compression intent.
For example, perceptual compression may employ perceptual quality metrics, whereas machine-vision compression may define $Q_{\mathcal{I}}$ according to the performance degradation of a downstream machine-vision task.
Here, $\mathbf{x}^{\star}$ serves only as an ideal reference for defining and evaluating the desired reconstruction quality and is not assumed to be available during compression procedure.

\subsection{Practical Image Coding Agents}
To realize the optimization in Eq.~\ref{eq:form}, we implement PICA as a practical multi-agent system that instantiates the agentic coding function $\mathcal F(\mathbf{x},\mathcal C,\mathcal I)$.
As illustrated in Fig.~\ref{fig:framework}, the \emph{Perception Agent} takes the observed image $\mathbf{x}=\mathcal G_{\delta}(\mathbf{x}^{\star})$ and user instruction $\mathcal I$ as inputs.
Using VLMs, it generates a structured configuration $\mathcal{S}$ specifying the instruction-dependent quality criterion $Q_{\mathcal{I}}(\cdot,\cdot)$ and the QR preference $\lambda_{\mathcal{I}}$ inferred from compression requirements such as the target bitrate.
The configuration also contains an input quality assessment, including the diagnosed degradation $\mathcal{G}'_{\delta}(\cdot)$, and auxiliary information such as image content analysis and identified ROIs.

Taking the observed image $\mathbf{x}$ and configuration $\mathcal S$ as inputs, the \emph{Improvement Agent} updates $\mathbf{x}$ through QR--aware restoration.
It encodes the image using the target codec $\mathcal C$ and applies SGQR-based reflection to improve the QR trade-off in subsequent coding iterations.
The \emph{Coding Agent} then encodes the updated image $\mathbf{x}$ using an off-the-shelf black-box codec $\mathcal C$ under the configuration $\mathcal S$, producing an aggregate coded representation $\mathcal B$ and its decoded reconstruction $\widetilde{\mathbf{x}}$.
It first performs a global QC search by initializing a codec-specific quality-control parameter and iteratively refining it based on the measured bitrate and reconstruction quality to meet the specified requirements.
To account for spatial variations in content importance under the target compression scenario $\mathcal I$, the Coding Agent further employs AROIC to preserve important regions while reducing the coding cost of less critical content.
Through SGQR-based plan reflection, AROIC enables region-adaptive compression for the objective, without codec-specific architectural modifications or retraining.
Finally, the \emph{After-Coding Improvement Agent} updates the decoded reconstruction $\widetilde{\mathbf{x}}$ using a restoration procedure similar to the Improvement Agent, but evaluates restoration candidates using image quality alone.
The resulting pair $(\mathcal B,\widetilde{\mathbf{x}})$ constitutes the final output of the agentic coding function $\mathcal F$ in Eq.~\ref{eq:form}.

\noindent\textbf{Scenario-Guided Quality--Rate Evaluation.}
To align candidate plan selection with the compression scenario specified by the user instruction $\mathcal{I}$, we propose an SGQR evaluation strategy.
Given a candidate improvement or coding plan, we obtain the resulting bitrate $b$ in bits per pixel and the corresponding decoded reconstruction $\widehat{\mathbf{x}}$.
Using an obtained reference image $\mathbf{x}_{\mathrm{ref}}$ for quality assessment, we define it as:
\begin{equation}
\mathcal{E}_{\mathrm{SGQR}}
\bigl(b,\widehat{\mathbf{x}},\mathbf{x}_{\mathrm{ref}};\mathcal{I}\bigr)
=
b+\lambda_{\mathcal{I}}
Q_{\mathcal{I}}
\bigl(\widehat{\mathbf{x}},\mathbf{x}_{\mathrm{ref}}\bigr).
\label{eq:sgqr}
\end{equation}
where $Q_{\mathcal{I}}(\widehat{\mathbf{x}},\mathbf{x}_{\mathrm{ref}})$ denotes the instruction-dependent reconstruction quality, with lower values indicating better quality.

In this work, we primarily consider three representative compression scenarios.
For \emph{standard compression}, $Q_{\mathcal{I}}$ is defined as the mean squared error between the reconstructed and reference images.
For \emph{perceptual compression}, it quantifies perceptual dissimilarity between these images using DISTS and LPIPS.
For \emph{machine-vision compression}, it is defined as the negative of a task-specific performance score evaluated on the reconstructed image (\eg, a weighted average of the product of detection confidence and IoU for object detection), with task predictions on the reference image serving as pseudo-ground truth.
The procedures for obtaining the bitrate $b$, reconstructed image $\widehat{\mathbf{x}}$, and reference image $\mathbf{x}_{\mathrm{ref}}$ in the Improvement and Coding Agents are detailed in Sec.~\ref{sec:improve} and Sec.~\ref{sec:coding}, respectively.

\subsection{Quality--Rate--Aware Improvement Agents}
\label{sec:improve}
We develop the \emph{Improvement Agent} and \emph{After-Coding Improvement Agent} to adaptively handle image degradations through restoration before encoding or after decoding for improved QR performance.

\noindent\textbf{Improvement Agent.}
Unlike previous restoration agents~\cite{zhu2025intelligent,zuo20254kagent} that primarily aim to improve image quality, our agent performs pre-coding restoration to improve the QR trade-off of subsequent compression.
For example, denoising can substantially reduce the bits spent on unwanted noise, but excessive denoising may remove meaningful details and impair reconstruction quality.
Conversely, deblurring can improve image quality, but the recovered details may require additional bits to encode.
These competing effects motivate jointly considering reconstructed image quality and bitrate when deciding which restoration operations to apply.
Guided by the diagnosed degradations $\mathcal{G}'_{\delta}(\cdot)$, the agent follows a \textit{scheduling--execution--reflection} pipeline.
It first schedules restoration subtasks, such as denoising or deblurring.
For each subtask, it applies $T-1$ candidate restoration tools to the subtask input and includes the unchanged input as a no-restoration baseline, producing $T$ uncompressed candidate images ${\mathbf{x}^{(t)}}_{t\in\mathcal{T}_{\mathrm{pre}}}$, where $\mathcal{T}_{\mathrm{pre}}=\{1,\ldots,T\}$.
It then evaluates the compression outcomes of these candidates and uses the resulting feedback to refine the remaining restoration plan.

Specifically, our reflection is \emph{QR-aware} and follows the SGQR criterion in Eq.~\ref{eq:sgqr}.
As noted in Sec.~\ref{sec:formulation}, the clean ground-truth image $\mathbf{x}^{\star}$ is unavailable during compression.
We therefore select the uncompressed candidate with the best no-reference image quality assessment (IQA) score from $\{\mathbf{x}^{(t)}\}_{t\in\mathcal{T}_{\mathrm{pre}}}$ and use it as a pseudo-reference, denoted by $\mathbf{x}^{(t_{\mathrm{ref}})}$.
Each candidate is then encoded and decoded using the target codec $\mathcal{C}$ at a common probe quality-control setting $q$ estimated from the QR preference $\lambda_{\mathcal I}$, producing a bitstream $\mathbf{b}_{\mathrm p}^{(t)}$ and a reconstruction $\widehat{\mathbf{x}}_{\mathrm p}^{(t)}$.
The selection procedure is formulated as:
\\
\begin{equation}
    \begin{aligned}
        & t^{\star}
        =
        \arg\min_{t\in\mathcal{T}_{\mathrm{pre}}}
        \mathcal{E}_{\mathrm{SGQR}}
        \!\left(
            R(\mathbf{b}_{\mathrm p}^{(t)}),
            \widehat{\mathbf{x}}_{\mathrm p}^{(t)},
            \mathbf{x}^{(t_{\mathrm{ref}})};
            \mathcal{I}
        \right), \\
        \text{s.t.,}\quad
        \bigl(
            & \mathbf{b}_{\mathrm p}^{(t)},
            \widehat{\mathbf{x}}_{\mathrm p}^{(t)}
        \bigr)
        =
        \mathcal{C}
        \!\left(\mathbf{x}^{(t)};q\right),
        t_{\mathrm{ref}}
        =
        \arg\min_{t\in\mathcal{T}_{\mathrm{pre}}}
        \cQ_{\mathrm{IQA}}
        \!\left(\mathbf{x}^{(t)}\right),
    \end{aligned}
    \label{eq:improvement_rd}
\end{equation}

\begin{figure*}[t!]
    \centering
    \begin{subfigure}{\linewidth}
        \centering
        \includegraphics[width=\linewidth]{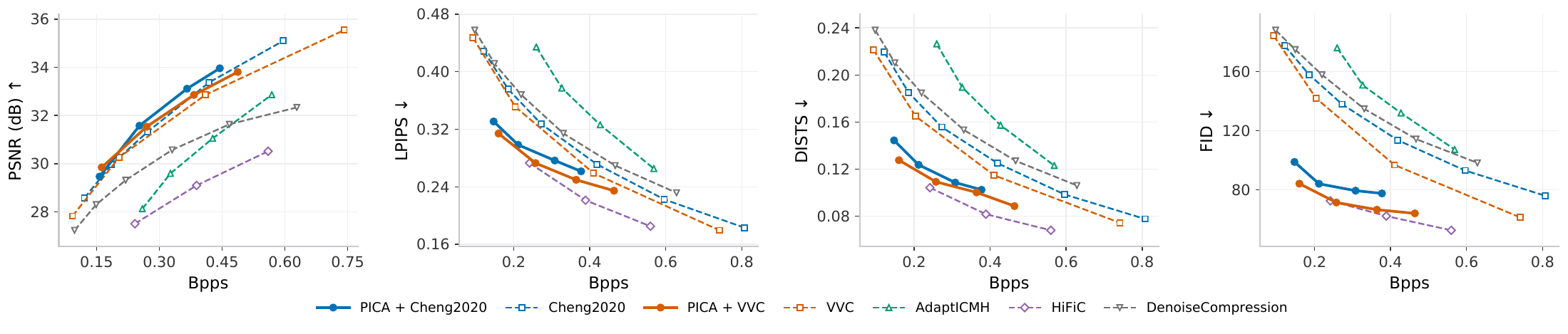}
        \caption{Quality-rate curves on the Kodak dataset.}
        \label{fig:sub1}
    \end{subfigure}

    \par\medskip

    \begin{subfigure}{\linewidth}
        \centering
        \includegraphics[width=\linewidth]{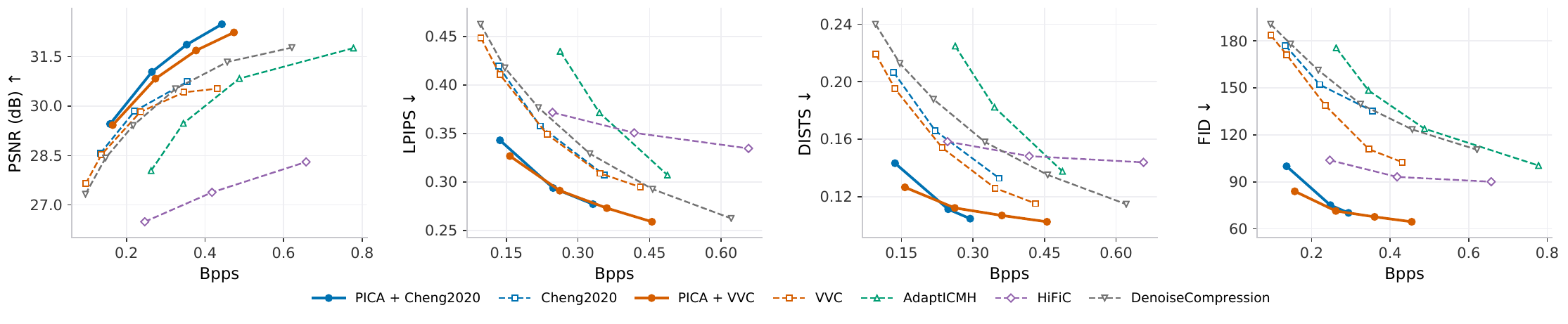}
        \caption{Quality-rate curves on the Noisy Kodak dataset.}
        \label{fig:sub2}
    \end{subfigure}

    \par\medskip

    \begin{subfigure}{\linewidth}
        \centering
        \includegraphics[width=\linewidth]{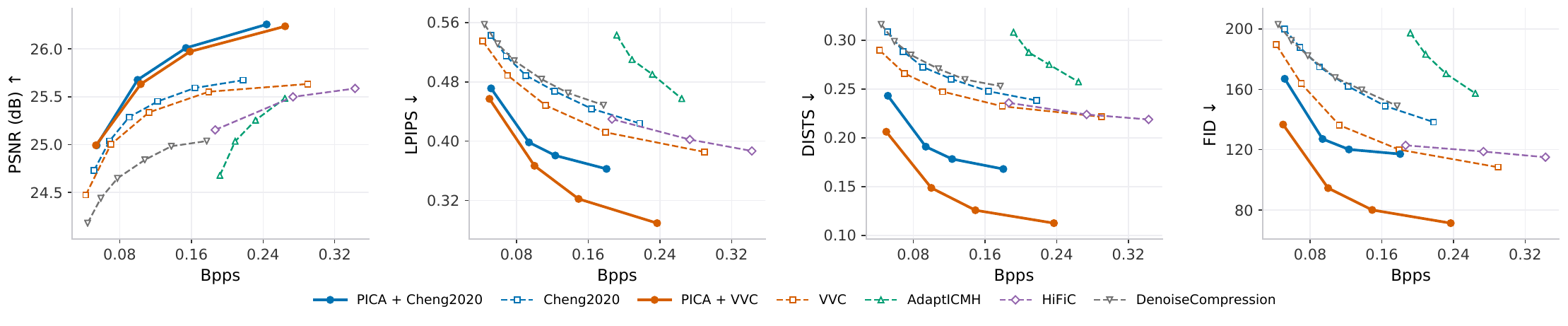}
        \caption{Quality-rate curves on the Blurred Kodak dataset.}
        \label{fig:sub3}
    \end{subfigure}
    \caption{Quality-rate performance on the (a) clean, (b) noisy, and (c) blurred Kodak benchmarks. Bitrate is measured in bits per pixel (bpps), while reconstruction quality is evaluated against the original clean reference images.}
    \label{fig:rq_kodak}
\end{figure*}

where $\cQ_{\mathrm{IQA}}(\cdot)$ denotes a composite no-reference IQA score computed, following~\cite{you2024depicting, you2025depictqav2}, with lower values indicating better image quality.
The terms $R(\mathbf{b}_{\mathrm p}^{(t)})$ and $\widehat{\mathbf{x}}_{\mathrm p}^{(t)}$ denote the measured bitrate (in bits per pixel) and the decoded reconstruction of the restored candidate $\mathbf{x}^{(t)}$, respectively, both obtained using the target codec $\mathcal{C}$.
The pseudo-reference $\mathbf{x}^{(t_{\mathrm{ref}})}$ is held fixed when comparing candidates within the current subtask, allowing the SGQR evaluation to assess reconstruction quality against a shared reference according to the specific compression scenario $\mathcal{I}$.

For each subtask, the agent selects the candidate $\mathbf{x}^{(t^{\star})}$ with the lowest estimated SGQR cost and updates $\mathbf{x}\leftarrow\mathbf{x}^{(t^{\star})}$ only if this cost is lower than that of the current image.
Otherwise, it retains the current image $\mathbf{x}$ and refines the remaining restoration plan based on agent feedback. Once restoration is complete, the agent passes the preprocessed image $\mathbf{x}$ to the Coding Agent for the encoding procedure.

\noindent\textbf{After-Coding Improvement Agent.}
This agent addresses potential residual degradations and compression-induced artifacts after decoding.
It follows a pipeline nearly identical to that of the Improvement Agent, with two main differences.
First, it uses a VLM to reassess the reconstructed image $\widetilde{\mathbf{x}}$ produced by the Coding Agent and identify residual degradations and artifacts to guide restoration planning.
Second, its reflection considers image quality alone using IQA, following~\cite{zhu2025intelligent,zuo20254kagent}:
\begin{equation}
    t^{\star}
    =
    \arg\min_{t\in\mathcal{T}_{\mathrm{post}}}
    \cQ_{\mathrm{IQA}}
    \!\left(\widetilde{\mathbf{x}}^{(t)}\right),
    \label{eq:after_coding_improvement}
\end{equation}
where $\widetilde{\mathbf{x}}^{(t)}$ denotes a candidate obtained by restoring the reconstructed image, and $\mathcal{T}_{\mathrm{post}}$ indexes the candidates for the current subtask, including the unchanged input as a no-restoration baseline.
Here, $\cQ_{\mathrm{IQA}}(\cdot)$ is the same no-reference IQA cost defined in Eq.~\ref{eq:improvement_rd}, with lower values indicating better image quality.
This IQA cost can be evaluated without access to the clean ground-truth image $\mathbf{x}^{\star}$.
% For machine-vision compression, reflection instead uses the task-specific quality measure $\cQ_I$, with task predictions on $\mathbf{x}_{\mathrm{ref}}$ serving as pseudo-ground truth.

For each subtask, the agent selects the candidate $\widetilde{\mathbf{x}}^{(t^{\star})}$ with the lowest IQA cost and updates $\widetilde{\mathbf{x}}\leftarrow\widetilde{\mathbf{x}}^{(t^{\star})}$ only if this cost is lower than that of the current image.
Otherwise, it retains $\widetilde{\mathbf{x}}$ and refines the remaining restoration plan based on agent feedback.
Once restoration is complete, the agent outputs the final reconstruction $\widetilde{\mathbf{x}}$.

% 正文
\begin{figure*}[t!]
    \centering
    \begin{subfigure}{\linewidth}
        \centering
        \includegraphics[width=\linewidth]{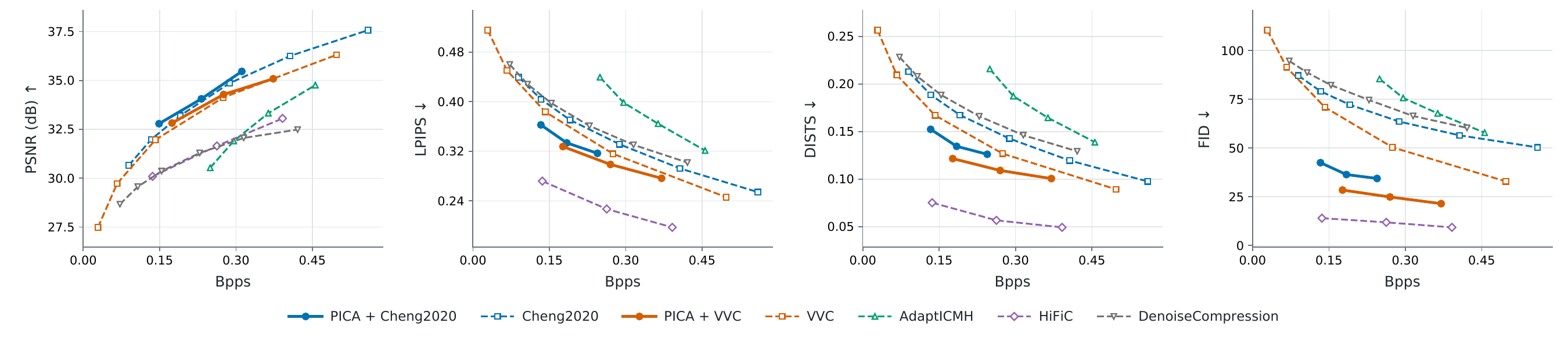}
        \caption{Quality-rate curves on the CLIC2020 dataset.}
        \label{fig:sub1}
    \end{subfigure}

    \par\medskip

    \begin{subfigure}{\linewidth}
        \centering
        \includegraphics[width=\linewidth]{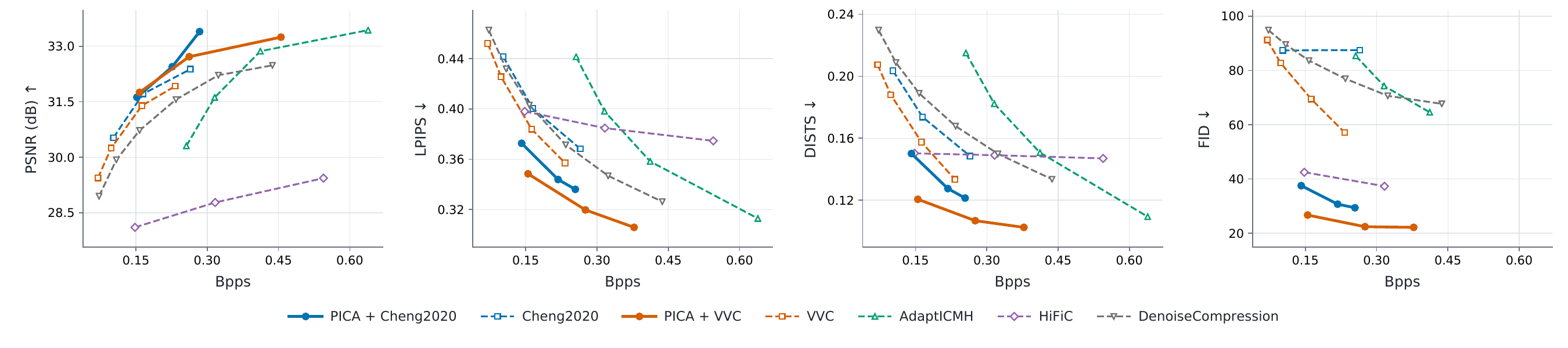}
        \caption{Quality-rate curves on the Noisy CLIC2020 dataset.}
        \label{fig:sub2}
    \end{subfigure}

    \par\medskip

    \begin{subfigure}{\linewidth}
        \centering
        \includegraphics[width=\linewidth]{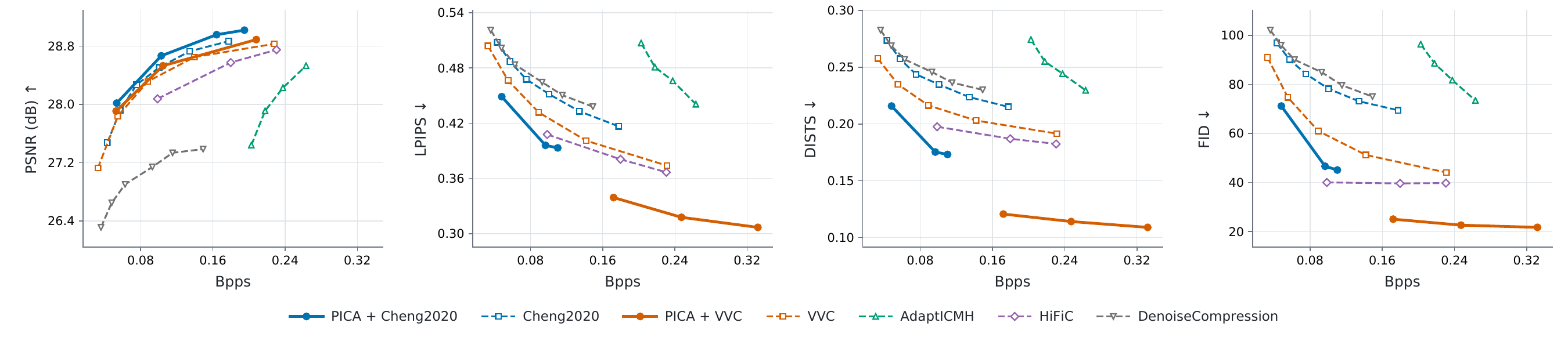}
        \caption{Quality-rate curves on the Blurred CLIC2020 dataset.}
        \label{fig:sub3}
    \end{subfigure}
    \caption{Quality-rate performance on the (a) clean, (b) noisy, and (c) blurred CLIC2020 benchmarks. Bitrate is measured in bits per pixel (bpps), while reconstruction quality is evaluated against the original clean reference images.}
    \label{fig:rq_clic}
\end{figure*}

\subsection{Codec-Agnostic Coding Agent}
\label{sec:coding}
Given the QR preference $\lambda_{\mathcal I}$, the compression requirements specified by $\mathcal I$, and the updated image $\mathbf{x}$ returned by the Improvement Agent, the {Coding Agent} searches for an appropriate coding configuration of the target black-box codec $\mathcal C$ to satisfy these requirements while balancing bitrate and reconstruction quality.
It employs two codec-agnostic mechanisms: global QC search and AROIC.

\noindent\textbf{Global QC Search.}
This mechanism identifies an appropriate global QC setting for the target codec $\mathcal C$ through feedback-driven refinement.
As illustrated in Fig.~\ref{fig:coding}, the agent initializes a global QC parameter $q_g$ based on the QR preference $\lambda_{\mathcal{I}}$, explicit compression requirements (\eg, target bitrate and compression ratio), and codec-specific QC knowledge (\eg, guidance on QP usage in H.266/VVC) provided in the configuration $\mathcal{S}$.
The codec then encodes and decodes $\mathbf{x}$ using $q_g$, and the measured bitrate and reconstruction quality are checked against the specified rate requirements and quality thresholds.
If these requirements are satisfied, the current setting $q_g$ is retained.
Otherwise, the agent generates feedback identifying the unmet requirements and updates $q_g\leftarrow q_g^{\mathrm{next}}$ accordingly, aiming to improve reconstruction quality when it is unacceptable or bring the bitrate closer to the target while maintaining acceptable quality.
This process continues until both the rate and quality requirements are satisfied or the iteration limit is reached.
The resulting global QC parameter $q_g$ is used in the subsequent AROIC procedure.
\begin{figure}[t!]
    \centering
    \begin{subfigure}{\linewidth}
        \centering
        \includegraphics[width=\linewidth, height=3.5cm]{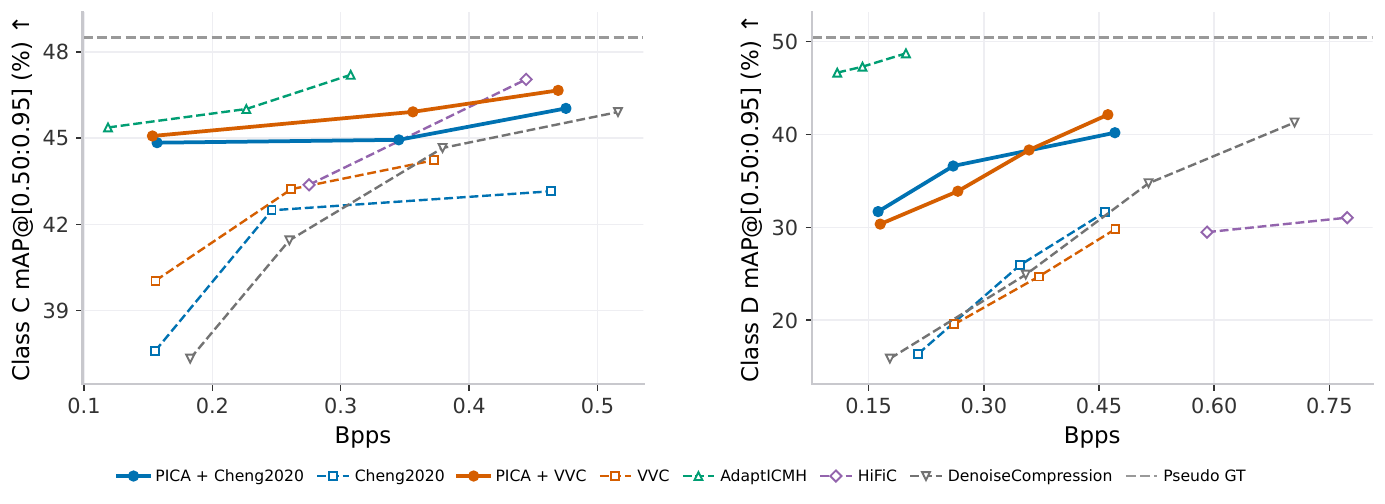}
        \caption{Quality-rate curves on the HEVC C and D datasets}
        \label{fig:sub1}
    \end{subfigure}

    \par\medskip

    \begin{subfigure}{\linewidth}
        \centering
        \includegraphics[width=\linewidth, height=3.5cm]{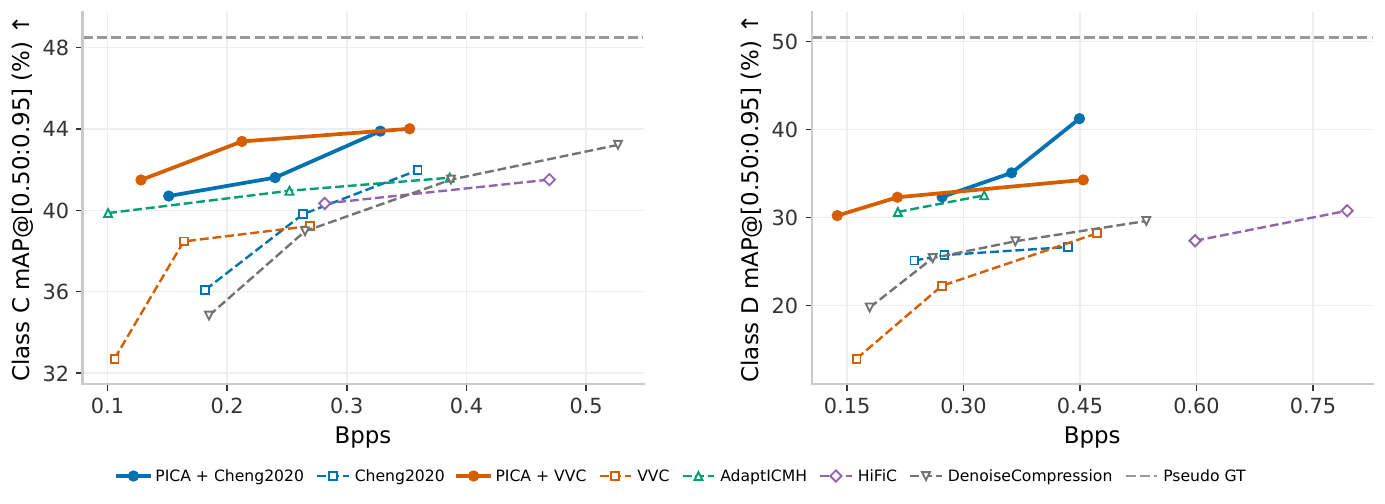}
        \caption{Quality-rate curves on the Noisy HEVC C and D datasets}
        \label{fig:sub2}
    \end{subfigure}

    \par\medskip
    
    \begin{subfigure}{\linewidth}
        \centering
        \includegraphics[width=\linewidth, height=3.5cm]{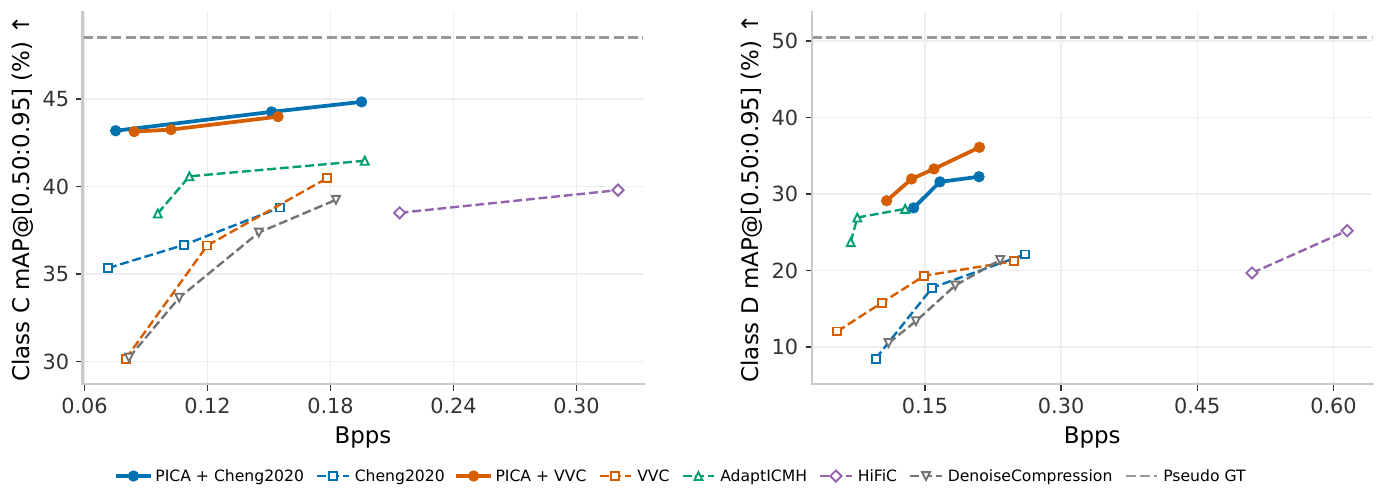}
        \caption{Quality-rate curves on the Blurred HEVC C and D datasets}
        \label{fig:sub3}
    \end{subfigure}

    \caption{Quality--rate performance on the (a) clean, (b) noisy, and (c) blurred benchmarks derived from HEVC Classes C and D. Bitrate is measured in bits per pixel (bpp), while object detection performance is evaluated on the reconstructed images using $\mathrm{mAP}_{[0.50:0.95]}$ (\%). The gray dashed lines indicate detection performance on uncompressed clean inputs.}
    \label{fig:rq_hevc}

\end{figure}

\begin{figure}[t!]
    \centering
    \begin{subfigure}{\linewidth}
        \centering
        \includegraphics[width=\linewidth, height=3.5cm]{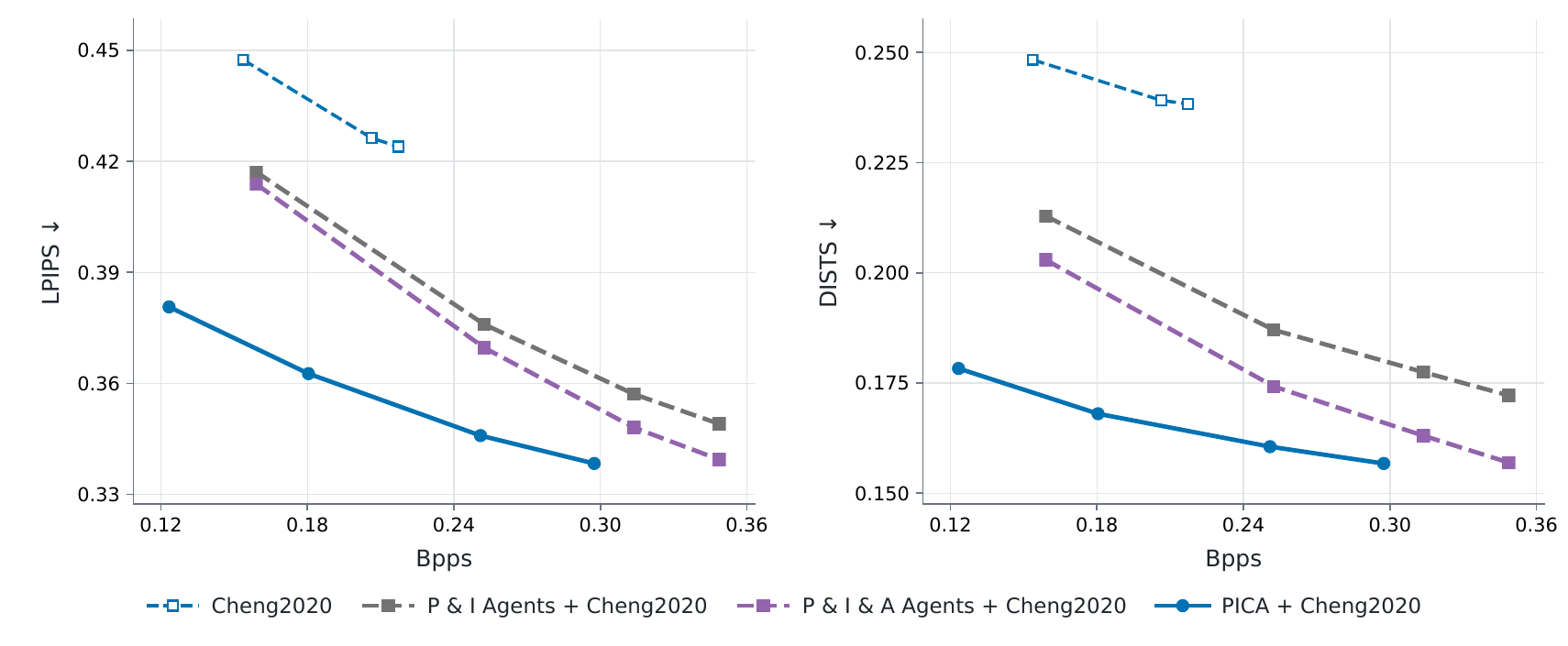}
        \caption{Effectiveness of each agent with Cheng2020 on Blurred Kodak.}
        \label{fig:sub1}
    \end{subfigure}

    \par\medskip

    \begin{subfigure}{\linewidth}
        \centering
        \includegraphics[width=\linewidth, height=3.5cm]{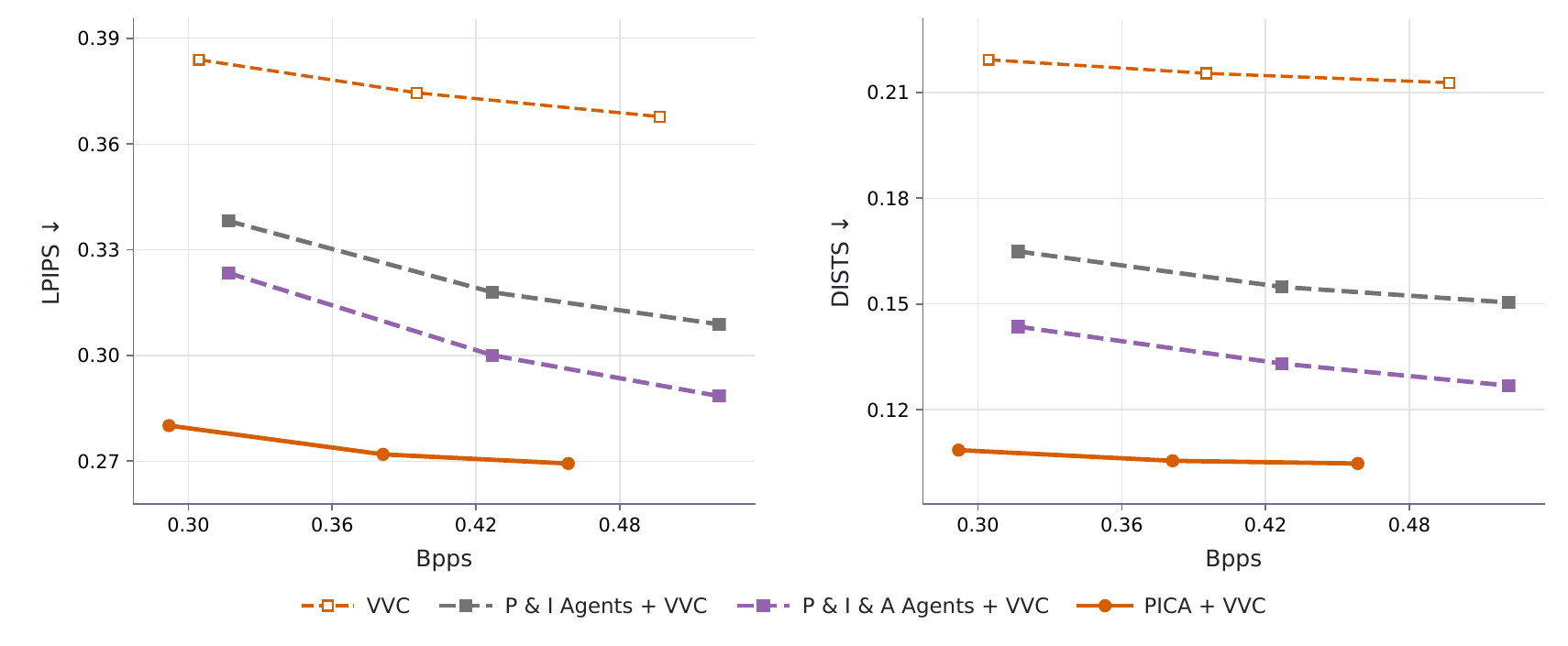}
        \caption{Effectiveness of each agent with VVC on Blurred Kodak.}
        \label{fig:sub2}
    \end{subfigure}

    \par\medskip
    
        \begin{subfigure}{\linewidth}
        \centering
        \includegraphics[width=\linewidth, height=3.5cm]{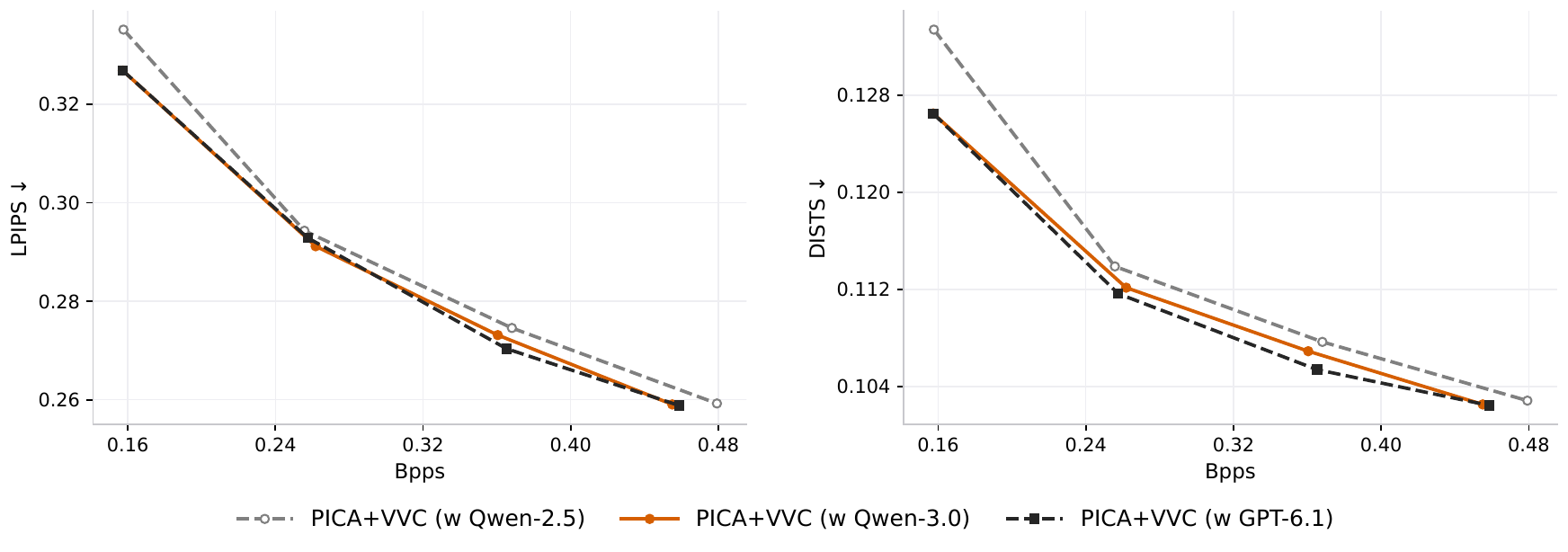}
        \caption{Robustness Across VLM Agent Backbones on Noisy Kodak.}
        \label{fig:sub3}
    \end{subfigure}

    \caption{Ablation studies of (a)–(b) the effectiveness of individual agents on the blurred Kodak dataset, where P, I, C, and A denote the Perception, Improvement, Coding, and After-coding Improvement Agents, respectively, and (c) PICA with different VLMs as agent backbones on the noisy Kodak dataset.}
    \label{fig:ablation}
\end{figure}

\noindent\textbf{Agentic ROI Coding.}
The relative importance of image regions varies across compression scenarios, making ROI coding an effective mechanism for allocating bits according to content importance.
However, reliance on codec-internal mechanisms~\cite{gao2025compx} can limit the transferability of conventional ROI coding methods across heterogeneous codecs.
To enable codec-agnostic regional adaptation, AROIC treats the target codec $\mathcal C$ as a black box and delegates regional coding decisions, execution, and refinement to the Coding Agent, without requiring codec-specific architectural modifications or retraining.
As illustrated in Fig.~\ref{fig:coding}, AROIC follows a \textit{scheduling--execution--reflection} workflow.
Given the preprocessed image $\mathbf{x}$ returned by the Improvement Agent and the $N$ ROI coordinates $\{\Omega_i\}_{i=1}^{N}$ provided by the Perception Agent, AROIC separates $\mathbf{x}$ into ROI crops and a background image:
\begin{equation}
\begin{aligned}
\mathbf{x}_{\mathrm{roi},i}
&= \operatorname{Crop}(\mathbf{x};\Omega_i),
\quad i=1,\ldots,N,\\
\mathbf{x}_{\mathrm{bg}}
&= (\mathbf{1}-\mathbf M)\odot\mathbf{x},
\end{aligned}
\label{eq:aroic_decomposition}
\end{equation}
where $\mathbf M$ is the binary mask covering the union of the ROIs and $\odot$ denotes element-wise multiplication.

First, the agent initializes the ROI and background QC settings from the global QC setting $q_g$ and \textit{schedules} $T-1$ candidate coding plans conditioned on the configuration $\mathcal S$, including the QR preference, compression requirements, image caption, and ROI information.
These plans adjust the background QC, ROI QC, or both to explore different QR trade-offs, and may additionally downsample the background before encoding to reduce its coding cost.

Then, the agent \textit{executes} each plan $t$ by invoking the target codec to independently encode and decode the background and the $N$ ROI crops using their respective QC settings, producing $N+1$ bitstreams and corresponding component reconstructions.
When background downsampling is used, the decoded background is upsampled to the original image resolution before the decoded ROI crops are placed at their corresponding spatial locations.
The resulting aggregate coded representation $\mathcal B^{(t)}$ and full-image reconstruction $\widehat{\mathbf{x}}^{(t)}$ are
\begin{equation}
\begin{aligned}
\mathcal B^{(t)}
&=
\left(
\mathbf b_{\mathrm{bg}}^{(t)},
\{\mathbf b_{\mathrm{roi},i}^{(t)}\}_{i=1}^{N},
\mathbf m^{(t)}
\right),\\
\widehat{\mathbf{x}}^{(t)}
&=
\operatorname{Agg}\!\left(
\widehat{\mathbf{x}}_{\mathrm{bg}}^{(t)},
\{\widehat{\mathbf{x}}_{\mathrm{roi},i}^{(t)}\}_{i=1}^{N};
\mathbf m^{(t)}
\right),
\end{aligned}
\label{eq:aroic_aggregation}
\end{equation}
where $\mathbf b_{\mathrm{bg}}^{(t)}$ and $\mathbf b_{\mathrm{roi},i}^{(t)}$ denote the background and $i$-th ROI bitstreams, respectively, and $\mathbf m^{(t)}$ contains the metadata required for reconstruction, including ROI locations and background scaling information.
The operator $\operatorname{Agg}$ spatially combines the background reconstruction $\widehat{\mathbf{x}}_{\mathrm{bg}}^{(t)}$ at the original resolution with the decoded ROI crops $\{\widehat{\mathbf{x}}_{\mathrm{roi},i}^{(t)}\}_{i=1}^{N}$ according to $\mathbf m^{(t)}$.

Next, the agent \textit{evaluates} the candidate coding plans indexed by $\mathcal T_{\mathrm{cod}}=\{1,\ldots,T\}$ using SGQR, with the uncompressed preprocessed image $\mathbf{x}$ serving as the shared quality reference and $R(\mathcal B^{(t)})$ denoting the total coding rate in bits per pixel. The candidate set also includes a global-QC baseline $(\mathcal B^{(T)},\widehat{\mathbf{x}}^{(T)})$, obtained by encoding the entire image (\ie, both ROIs and the background) at the unchanged global QC setting $q_g$. The best candidate is then selected as:
\begin{equation}
t^\star
=
\arg\min_{t\in\mathcal T_{\mathrm{cod}}}
\mathcal E_{\mathrm{SGQR}}
\!\left(
R(\mathcal B^{(t)}),
\widehat{\mathbf{x}}^{(t)},
\mathbf{x};
\mathcal I
\right).
\label{eq:aroic_selection}
\end{equation}

\begin{figure*}[t!]
    \centering
    \includegraphics[width=\linewidth]{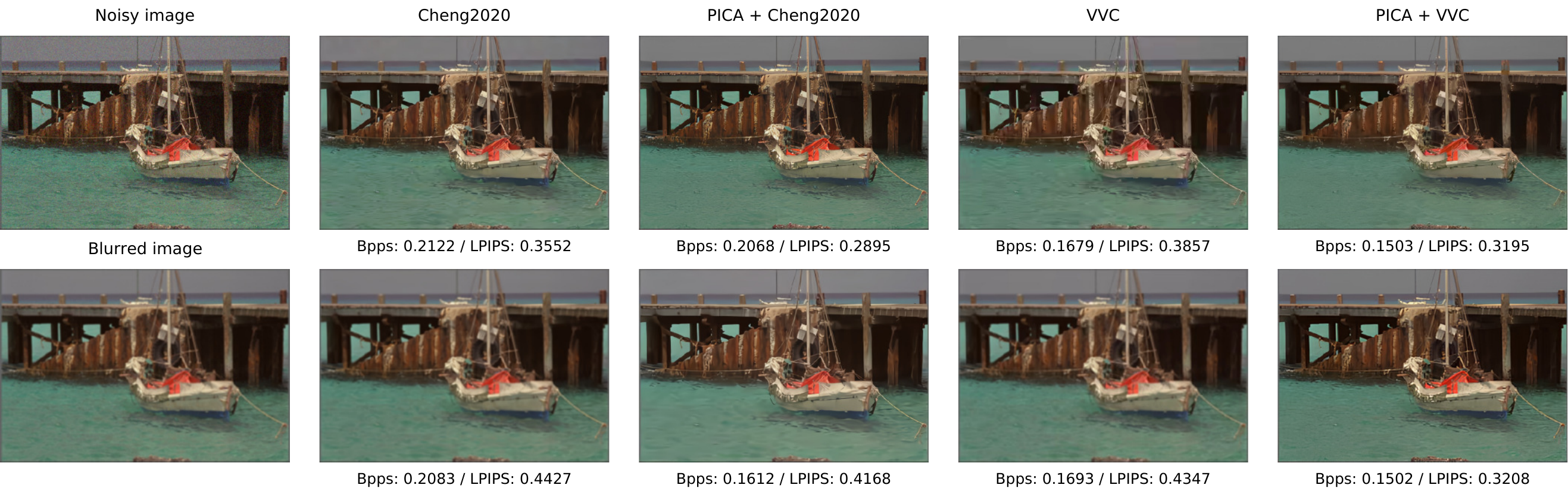}
    \includegraphics[width=\linewidth]{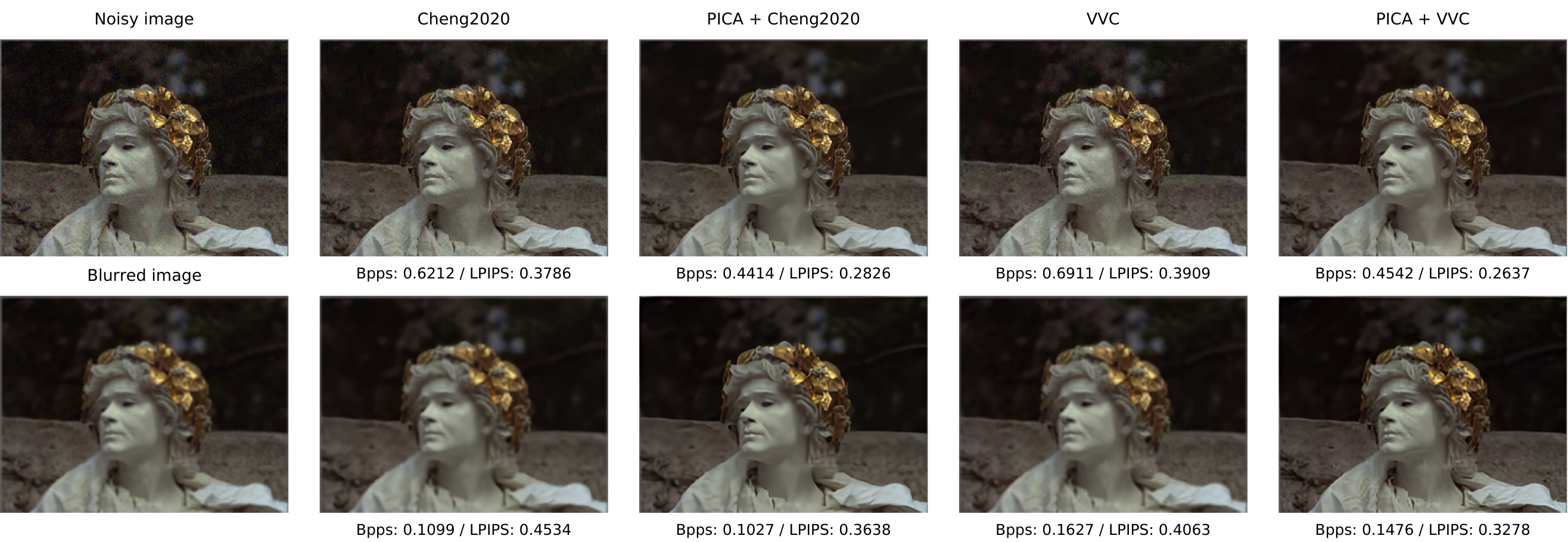}
    \includegraphics[width=\linewidth]{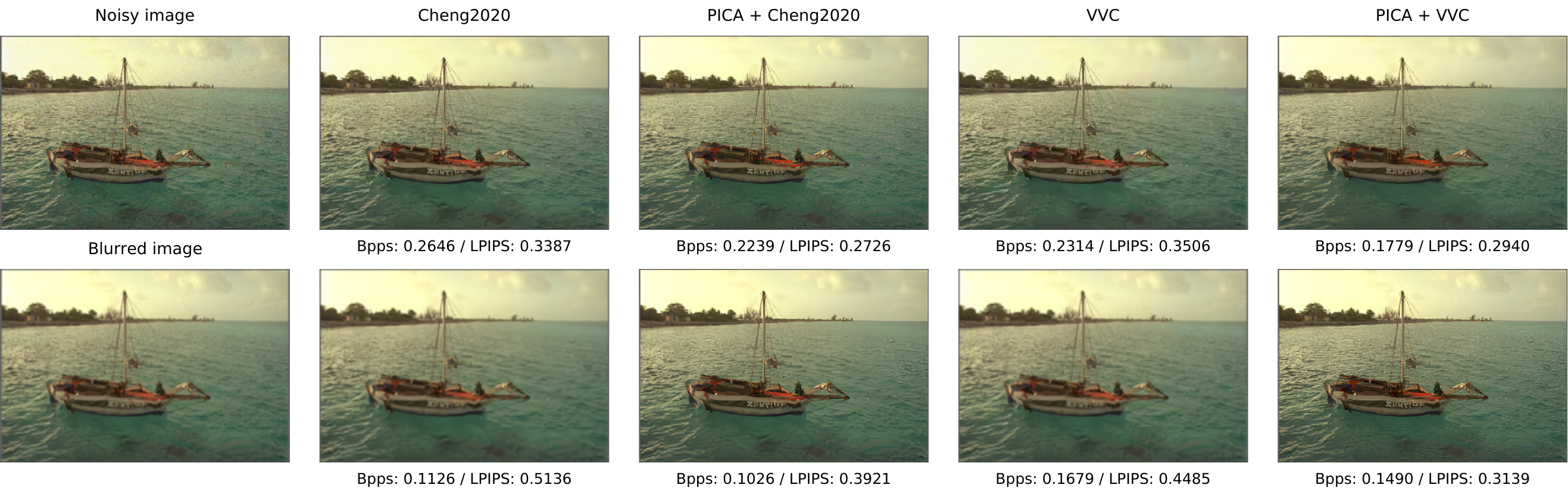}
    \includegraphics[width=\linewidth]{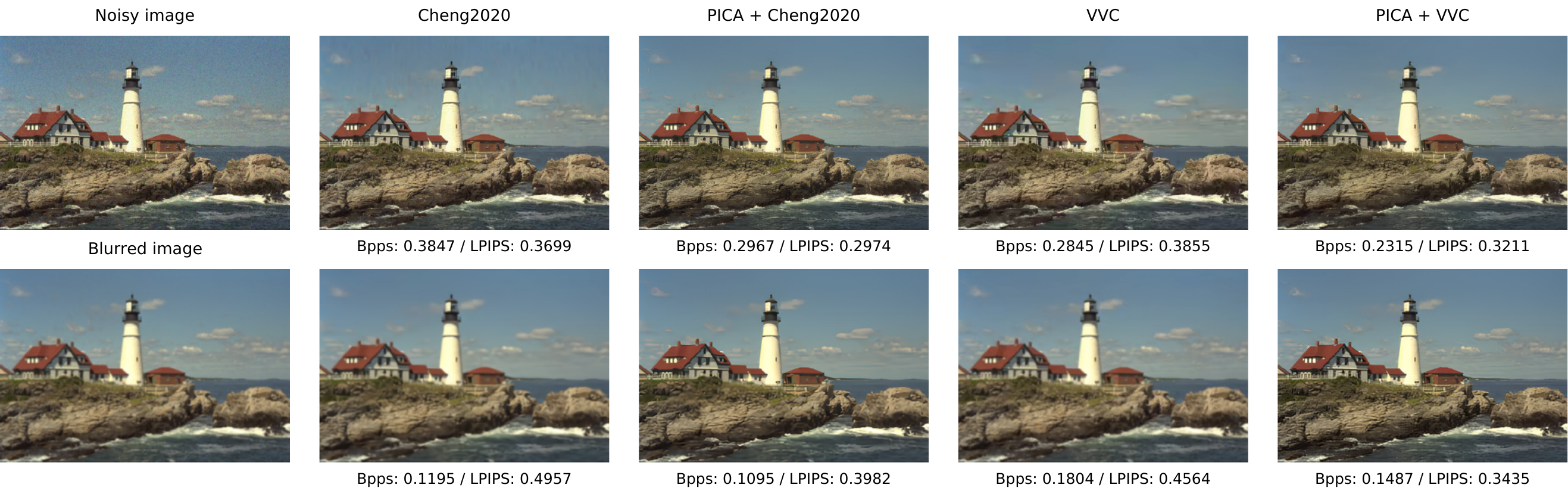}
    \caption{Qualitative comparison of PICA using VVC and Cheng2020 as base codecs against their corresponding baseline codecs.}
    \label{fig:vis}
\end{figure*}

Finally, the agent \textit{reflects} on the selected candidate $\widehat{\mathbf{x}}^{(t^\star)}$.
If the candidate achieves a lower SGQR cost than the global-QC baseline while satisfying both the quality and bitrate requirements, the agent updates $(\mathcal B,\widetilde{\mathbf{x}})\leftarrow(\mathcal B^{(t^\star)},\widehat{\mathbf{x}}^{(t^\star)})$.
Otherwise, it generates actionable feedback and \textit{reschedules} the coding plans accordingly.
This feedback guides further adjustments to ROI QC, background QC, or background downsampling, after which the revised plans are executed and evaluated again.
The process terminates when an QR improvement is obtained or the iteration limit is reached. AROIC then returns the selected coded representation $\mathcal B$ and reconstruction $\widetilde{\mathbf{x}}$, or the global-QC baseline if no candidate satisfies the acceptance criteria.

\subsection{Models and Tools}
\label{sec:models_tools}

We summarize the pretrained models and task-specific tools used in PICA and its evaluation below:

\noindent\textbf{Denoising:}
X-Restormer~\cite{chen2024xrestormer}, SwinIR~\cite{liang2021swinir}, MPRNet~\cite{zamir2021mprnet}, MAXIM~\cite{tu2022maxim}, Restormer~\cite{zamir2022restormer}, and NAFNet~\cite{chen2022nafnet}.

\noindent\textbf{Defocus Deblurring:}
DRBNet~\cite{ruan2022drbnet}, IFAN~\cite{lee2021ifan}, Restormer~\cite{zamir2022restormer}, Diff-Plugin~\cite{liu2024diffplugin}, and LaKDNet~\cite{ruan2023lakdnet}.

\noindent\textbf{Super-resolution:}
DiffBIR~\cite{lin2024diffbir}, X-Restormer~\cite{chen2024xrestormer}, SwinIR~\cite{liang2021swinir}, HAT~\cite{chen2023hat}, OSEDiff~\cite{wu2024osediff}, PiSA-SR~\cite{sun2025pisasr}, SwinFIR~\cite{zhang2022swinfir}, NAFSSR~\cite{chu2022nafssr}, and HMANet~\cite{chu2024hmanet}.

\noindent\textbf{Compression Artifact Removal:}
FBCNN~\cite{jiang2021fbcnn}, SwinIR~\cite{liang2021swinir}, WildIR~\cite{luo2024wildir} and NOLAIR~\cite{kim2026nolair}.

\noindent\textbf{Detection:}
YOLOv8~\cite{jocher2023yolov8} for object-based ROI detection and EasyOCR~\cite{jaidedaiEasyocr} for text-based ROI detection. Faster R-CNN~\cite{ren2015fasterrcnn} with a ResNeXt-101-32x8d-FPN backbone for machine-vision compression evaluation.

\noindent\textbf{IQA:}
CLIP-IQA+~\cite{wang2023clipiqa}, MANIQA~\cite{yang2022maniqa}, MUSIQ~\cite{ke2021musiq}, and NIQE~\cite{mittal2013niqe} for no-reference IQA, as~\cite{zuo20254kagent, zhu2025intelligent}.
LPIPS~\cite{zhang2018lpips} and DISTS~\cite{ding2022dists} for full-reference perceptual comparisons.

\noindent\textbf{VLMs:}
DepictQA-family model~\cite{you2024depictqa,you2025depictqav2} for qualitative IQA, including degradation identification and severity estimation.
Qwen3-VL-8B-Instruct~\cite{bai2025qwen3} supports user instruction interpretation, image analysis, task planning, tool invocation, and feedback-guided plan refinement across all agents.

\section{Experiments}
\subsection{Datasets and Protocols.}
\noindent \textbf{Datasets.}
We evaluate our methods on three datasets: Kodak, CLIC2020, and HEVC. Kodak~\cite{franzen1999kodak} consists of 24 natural images and is a widely used benchmark for image compression. The CLIC2020 validation set~\cite{toderici2020clic} contains 102 high-resolution images spanning diverse scenes and textures, enabling evaluation on more varied visual content.
For machine-vision compression, we use images sampled from the HEVC test sequences~\cite{sullivan2012overview}, with ground-truth object categories and bounding boxes provided by SFU-HW-Objects-v3~\cite{choi2021dataset, choi2024sfuhw}. Specifically, we select 8 sequences from Classes C and D, divide each sequence into 6 equal temporal segments, and extract the $1^{st}$ frame from each segment, yielding 48 images.

\noindent \textbf{Degradation} 
We consider two common image degradations that can substantially affect compression performance, which are noise and blur~\cite{cheng2022optimizing,brummer2023importance,ye2023accelir,10.1145/3743093.3770957}. Following prior agentic image restoration methods~\cite{zhu2025intelligent,zuo20254kagent}, we synthetically generate additive Gaussian noise and defocus blur from the original images. Specifically, we add Gaussian noise with a standard deviation of $\sigma=15$ in the 8-bit pixel intensity range. To generate defocus blur, we use a circular disk point-spread function with a radius of 3 pixels, smoothed with a $3\times3$ Gaussian filter, to blur each image. We apply each degradation independently to the images in the Kodak dataset, the CLIC2020 validation set, and the HEVC test set. The original images form the clean benchmarks. Thus, each dataset has 3 benchmarks: clean, Gaussian-noise, and defocus-blur. In total, there are 9 compression benchmarks: Kodak, Noisy Kodak, Blurred Kodak, CLIC2020, Noisy CLIC2020, Blurred CLIC2020, HEVC, Noisy HEVC, and Blurred HEVC.

\noindent\textbf{Evaluation Metrics.}
For the 6 clean, noise and blur benchmarks derived from the Kodak and CLIC2020 validation sets, we evaluate reconstruction fidelity using PSNR, the most widely used objective metric for image compression. We also report 3 perceptual metrics: LPIPS and DISTS, computed per image and averaged over each benchmark, and FID, computed at the domain level by comparing the distributions of reconstructed images and their clean references. Here, clean references denote the original ground-truth images without synthetic degradation. For machine-vision compression, we evaluate object detection performance using $\mathrm{mAP}_{[0.50:0.95]}$, following~\cite{liu2026eac}. Faster R-CNN~\cite{ren2015fasterrcnn} with a ResNeXt-101-32x8d-FPN backbone is used as the downstream detector for all methods. All bitrates are reported in bits per pixel (Bpps), calculated by dividing the number of coded bits by the number of pixels in the corresponding image.

\subsection{Baseline Codecs.}
We evaluate our PICA using two representative codecs: the conventional H.266/VVC codec in intra-frame coding mode (\ie,  VVC)~\cite{bross2021vvc} and the neural image codec Cheng2020~\cite{cheng2020compression}. For VVC, we use the VTM-24.0 reference software~\cite{vtm24} with \texttt{encoder\_intra\_vtm.cfg} and the reduced-runtime configuration. For Cheng2020, we use the \texttt{Cheng2020Anchor} variant implemented in CompressAI-1.2.8~\cite{begaint2020compressai}. For each codec, we assume only that the agent has high-level prior knowledge of how a single control parameter affects bitrate and reconstruction quality. That is the quantization parameter (\ie, QP) for VVC and the quality level (\ie,  q) for Cheng2020. Our PICA is training-free and requires no knowledge of codec internals or additional codec-specific expertise

In addition to the two base codecs for standard image compression, we include three specialized codecs designed and optimized for specific compression scenarios: HiFiC~\cite{mentzer2020high} for perceptual compression, Adapt-ICMH~\cite{li2024adapticmh} for machine-vision compression, and DenoiseCompression~\cite{cheng2022optimizing} for compression of noisy images. We use an open-source PyTorch reimplementation~\cite{han2020hificpytorch} of HiFiC and the official open-source implementations of Adapt-ICMH and DenoiseCompression.

% PICA is training-free and requires no knowledge of codec internals or additional codec-specific expertise.

% These BD-rate savings are computed over overlapping quality ranges and averaged only over valid benchmarks, with six, six, and three valid benchmarks for Cheng2020 and four, three, and two for VVC, respectively.
\subsection{Experiment Results.}
We evaluate compression performance across different compression scenarios and input conditions on Kodak, CLIC2020, and HEVC in Fig.~\ref{fig:rq_kodak}, Fig.~\ref{fig:rq_clic}, and Fig.~\ref{fig:rq_hevc}, respectively.
In general, PICA consistently improves the QR performance of both baseline codecs across the evaluated scenarios and degradations, demonstrating its ability to adapt Cheng2020 and VVC to diverse compression objectives.
For example, in perceptual compression, PICA achieves average bitrate savings at matched LPIPS of 34.52\%, 44.90\%, and 62.66\% over Cheng2020, and 28.96\%, 43.17\%, and 59.95\% over VVC, on the clean, noisy, and blurred Kodak benchmarks.
For machine-vision compression on HEVC Class D, PICA achieves average relative gains in $\mathrm{mAP}_{[0.50:0.95]}$ of 52.58\%, 40.94\%, and 55.77\% over Cheng2020, and 62.28\%, 80.46\%, and 55.56\% over VVC, under clean, noisy, and blurred input conditions, respectively, at matched bitrates.
Notably, even though both baseline codecs are specifically optimized for standard compression, PICA achieves further average bitrate savings at matched PSNR of 8.96\%, 22.86\%, and 31.07\% over Cheng2020, and 7.49\%, 22.99\%, and 35.81\% over VVC, on the clean, noisy, and blurred Kodak benchmarks.

On the other hand, we observe that codecs optimized for specific compression objectives or input conditions often exhibit limited generalization across scenarios and degradations.
For instance, HiFiC achieves strong perceptual performance on clean images, but requires 185.89\% and 111.57\% more bitrate than PICA+VVC at matched PSNR for standard compression on clean Kodak and CLIC2020, respectively.
Its perceptual performance also deteriorates under input degradations.
On the degraded Kodak and CLIC2020 benchmarks, HiFiC exhibits substantially higher LPIPS scores than PICA+VVC at comparable bitrates, indicating much lower perceptual quality.
Similarly, DenoiseCompression, specifically optimized for standard compression of noisy images, exhibits limited robustness to blur.
Adapt-ICMH performs well in machine-vision compression on clean images but remains less competitive in standard and perceptual compression, with its machine-vision performance also falling behind PICA under degradations.

\subsection{Ablation Study}
\noindent\textbf{Effectiveness of Individual Agents.}
We evaluate the effectiveness of the agents on the blurred Kodak benchmark by progressively incorporating them into Cheng2020 and VVC.
Specifically, we introduce two intermediate variants: the baseline codec equipped with the Perception and Improvement Agents (\ie, P \& I), and a variant that additionally incorporates the After-coding Improvement Agent (\ie, P \& I \& A).
Fig.~\ref{fig:ablation}~(a) and (b) compare these variants with the baseline codecs and the full PICA framework.
% All percentage reductions below are averaged at matched bitrates over the overlapping bitrate range of each pair of curves.
% %
These results demonstrate the complementary benefits of pre-coding restoration, post-coding restoration, and AROIC.
Our observation is that, by enabling QR-aware pre-coding restoration, the P \& I variant reduces LPIPS by 6.87\% and 13.94\%, and DISTS by 15.83\% and 27.15\%, relative to Cheng2020 and VVC, respectively.
Adding the After-coding Improvement Agent for quality-only post-coding restoration further reduces LPIPS by 1.76\% and 5.52\%, and DISTS by 6.83\% and 14.07\%, relative to the corresponding P \& I variants, which use pre-coding restoration only.
The full PICA framework achieves the best QR performance among the evaluated variants.
Compared with the restoration-only P \& I \& A variants, incorporating the Coding Agent with AROIC further reduces LPIPS by 7.62\% and 11.56\%, and DISTS by 10.05\% and 22.66\%, for Cheng2020 and VVC, respectively.
These results demonstrate the complementary benefits of pre-coding restoration, post-coding restoration, and AROIC.

\noindent\textbf{Robustness Across VLM Agent Backbones.}
To evaluate PICA's robustness to the choice of VLM agent backbone, we compare three VLMs: two open-source models, Qwen2.5-VL-7B-Instruct~\cite{bai2025qwen25vl} and Qwen3-VL-8B-Instruct~\cite{bai2025qwen3}, and one commercial model, GPT-6.1-Sol~\cite{openai_gpt6}.
Here, the VLM agent backbone refers to the model that supports user instruction interpretation, image analysis, task planning, tool invocation, and feedback-guided plan refinement across all agents.
Fig.~\ref{fig:ablation} (c) presents the results on the noisy Kodak benchmark.
The three backbones yield nearly identical QR performance, suggesting that PICA's effectiveness is robust to the choice of backbone among the evaluated VLMs.
These results support flexible deployment with different VLM backbones, allowing the choice to accommodate available computational resources and deployment requirements.

\subsection{Qualitative Study}
In addition to the quantitative evaluation, Fig.~\ref{fig:vis} presents qualitative comparisons using four samples each from the noisy and blurred Kodak benchmarks.
We compare Cheng2020 and VVC with their PICA-enhanced counterparts.
Across both degradations, PICA produces reconstructions with better visual quality and lower LPIPS scores while reducing bitrate.

For noisy images, noise introduces unwanted image variations that impair visual quality and consume additional bits during encoding.
As shown in the $2^{nd}$ and $4^{th}$ columns, both the neural codec Cheng2020 and the traditional codec VVC retain visible residual noise in their reconstructed images.
This is particularly evident on the statue's face in the $3^{rd}$ row and in the sky in the $7^{th}$ row.
PICA suppresses this noise more effectively, improving perceptual quality while reducing the bits spent on encoding unwanted content.

For blurred images, blur substantially degrades fine details, as illustrated by the water surface textures in the sixth row and the rock textures in the eighth row.
The baseline codecs do not effectively restore these details, whereas PICA produces clearer and more discernible textures through deblurring.
Despite restoring texture details, PICA achieves lower LPIPS scores without increasing bitrate relative to the corresponding baselines.
These observations further support the effectiveness of our SGQR strategy in jointly considering bitrate and reconstruction quality when selecting restoration and coding plans, rather than pursuing restoration quality alone.

\section{Conclusion}

In this work, we introduced PICA, a training-free multi-agent framework that adapts existing image codecs to diverse compression scenarios and potentially degraded inputs.
It combines four collaborative agents with SGQR evaluation and feedback-guided reflection to coordinate restoration and coding according to specific user requirements and potential input degradations.
Extensive experiments on the Kodak, CLIC2020, and HEVC datasets demonstrate that PICA improves the quality--rate performance of both the conventional codec H.266/VVC and the neural codec Cheng2020 across standard, perceptual, and machine-vision compression under clean, noisy, and blurred input conditions.
These improvements are achieved without codec-specific architectural modifications, retraining on large-scale datasets, or detailed knowledge of codec internals.
By decoupling scenario adaptation from extensive codec redevelopment and scenario-specific training, PICA enables advances in standard codecs to benefit a broader range of image processing and transmission applications.
More broadly, this work connects agentic decision-making with image compression research and motivates further exploration of VLM-based agents for compression and related visual processing tasks.

% \section{Acknowledgment}
% This work was initiated and largely completed during Zhenghao Chen's participation in the Microsoft Research Asia (MSRA) StarTrack Scholars Program.

% \end{thebibliography}
{\small
\bibliographystyle{IEEEtran}
\bibliography{mybib}
}
\end{document}